\documentclass{article}

\PassOptionsToPackage{numbers, compress}{natbib}
\usepackage[eandd, final]{neurips_2026}

\usepackage{url}            
\usepackage[colorlinks=true,linkcolor=black,citecolor=black,urlcolor=blue]{hyperref}       
\usepackage{booktabs}       
\usepackage{arydshln}       
\usepackage{amsfonts}       
\usepackage{nicefrac}       
\usepackage{microtype}      
\usepackage[dvipsnames]{xcolor}         
\usepackage{colortbl}       
\usepackage{graphicx}       
\usepackage[most]{tcolorbox} 

\newcommand{\dashrule}{\noindent\makebox[\linewidth]{\color{black!40}\dashrulefill}}\newcommand{\dashrulefill}{\leaders\hbox to 5pt{\hss\rule{3pt}{0.4pt}\hss}\hfill}
\usepackage{enumitem}

\title{TransitLM: A Large-Scale Dataset and Benchmark for Map-Free Transit Route Generation}

\author{%
  Hanyu Guo\rlap{$^{*}$} \quad Jiedong Yang\rlap{$^{*}$} \quad Chao Chen \quad Longfei Xu \quad Kaikui Liu \quad Xiangxiang Chu \\[0.5em]
  Amap, Alibaba Group\\
  Beijing, China\\[0.3em]
}

\begin{document}

\renewcommand{\thefootnote}{*}
\footnotetext{Equal contribution.}
\renewcommand{\thefootnote}{\arabic{footnote}}

\maketitle

\begin{abstract}
    Public transit route planning traditionally depends on structured map infrastructure and complex routing engines, and no existing dataset supports training models to bypass this dependency. We present TransitLM, a large-scale dataset of nearly 13 million transit route planning records from four Chinese cities covering 120,845 stations and 13,666 lines, released as a continual pre-training corpus and benchmark data for three evaluation tasks with complementary metrics. Experiments show that an LLM trained on TransitLM produces structurally valid routes at high accuracy and implicitly grounds arbitrary GPS coordinates to appropriate stations without any explicit mapping. These results demonstrate that transit route planning can be learned entirely from data, enabling end-to-end, map-free route generation directly from origin-destination information. The dataset and benchmark are available at \url{https://huggingface.co/datasets/GD-ML/TransitLM}, with evaluation code at \url{https://github.com/HotTricker/TransitLM}.
\end{abstract}

\section{Introduction}
\label{sec:introduction}
Public transit route planning underpins daily urban mobility, yet conventional systems rely heavily on structured map infrastructure and complex engineering pipelines for candidate retrieval and ranking over topological networks. Massive route planning logs continuously generated by transit platforms implicitly encode rich routing knowledge, including boarding stations, transfer points, and how travelers balance speed, convenience, and line preference. This contrast motivates a natural question: can route planning be learned directly from such data, bypassing maps and routing engines entirely?

One might expect general-purpose LLMs like GPT-3 \cite{brown2020language}, GPT-4 \cite{openai2023gpt4}, and Qwen3 \cite{yang2025qwen3} to address this question with their strong reasoning and broad world knowledge. However, recent studies argue that autoregressive LLMs cannot reliably perform planning by themselves \cite{valmeekam2023planning,kambhampati2024llms}. Although these models may recall frequently mentioned stations or popular routes, they consistently produce routes with hallucinated stations or broken connections \cite{huang2023hallucination}, particularly for less prominent origin-destination pairs. This limitation stems from the absence of suitable training data. Existing data sources each capture only partial aspects of the problem. Vehicle trajectory datasets such as T-Drive \cite{yuan2010tdrive} and Porto Taxi \cite{moreira2013porto} lack station structures. Static network datasets including GTFS \cite{wong2013leveraging} and CPTOND-2025 \cite{wang2026china} contain no user behavior or planning trajectories. Consequently, no existing source provides the complete route structures and behavioral annotations needed for learning end-to-end transit planning.

\begin{figure}[t]
\centering
\includegraphics[width=0.96\textwidth]{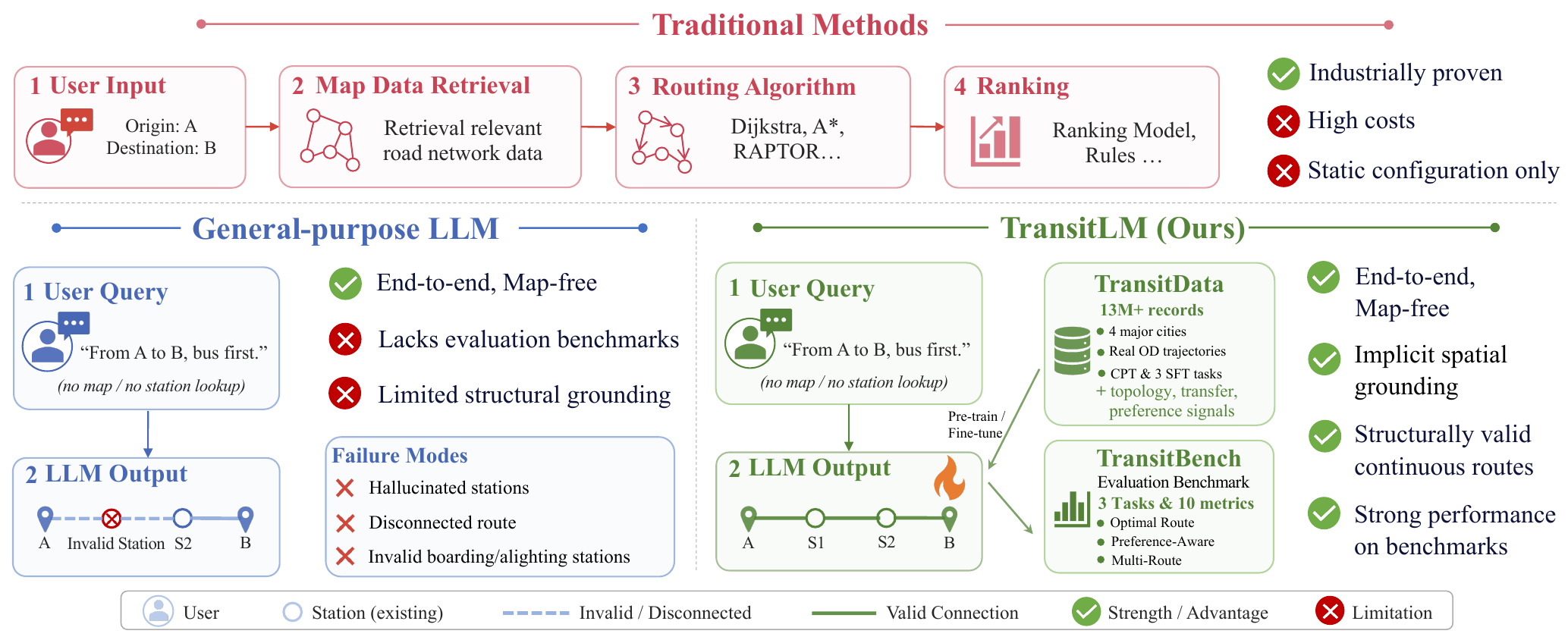}
\caption{Three paradigms for transit route planning. \textbf{Top:} Traditional map-based pipeline. \textbf{Bottom-left:} General-purpose LLMs lack structural grounding, producing hallucinated stations, disconnected routes, and invalid boarding/alighting points. \textbf{Bottom-right:} TransitLM generates structurally valid, continuous routes end-to-end via implicit spatial grounding, without map infrastructure.}
\label{fig:method_comparison}
\end{figure}

As illustrated in Figure~\ref{fig:method_comparison}, we introduce TransitLM to address this gap. TransitLM is a large-scale dataset of nearly 13 million route planning records from four Chinese cities: Beijing, Shanghai, Shenzhen, and Chengdu, covering 120,845 stations and 13,666 transit lines. Each record captures a full planning session with GPS coordinates, station sequences, transfer points, line identifiers, segment-level timing, and route-type annotations. We release two complementary resources. The continual pre-training corpus contains textual route descriptions for next-token prediction training, enabling models to internalize transit network topology and spatial relationships. The benchmark-specific SFT data provides standardized prompts and labels for three core tasks: optimal route generation, preference-aware planning, and multi-route generation, each evaluated by complementary metrics spanning connectivity, access feasibility, route overlap, and numeric field accuracy.

To validate the dataset, we train an LLM through continual pre-training followed by supervised fine-tuning. Our experiments reveal three key findings.
\textbf{(1)~End-to-end map-free route generation is feasible.} The trained model produces structurally valid, connected routes at high accuracy, demonstrating that rich trajectory data alone can match conventional map-based routing engines.
\textbf{(2)~Implicit spatial grounding emerges from data.} Given only origin and destination GPS coordinates, the model learns to resolve arbitrary coordinates to appropriate boarding and alighting stations without any explicit coordinate-to-station mapping or geographic database, effectively internalizing the spatial topology of the transit network.
\textbf{(3)~A single model generalizes across planning objectives.} A jointly trained model matches or exceeds task-specific counterparts on all three benchmarks without negative transfer, confirming that the transit knowledge encoded in the dataset is task-agnostic and supports unified deployment across diverse planning scenarios.
Our contributions are as follows:
\begin{itemize}[left=0pt]
\item \textbf{Dataset.} We present TransitLM, a large-scale dataset of nearly 13 million transit route planning records spanning four Chinese cities, 120,845 stations, and 13,666 lines, released as a pre-training corpus and benchmark data with standardized prompts and labels.
\item \textbf{Benchmark.} We define three evaluation tasks: optimal route generation, preference-aware planning, and multi-route generation. Each task is evaluated by complementary metrics spanning connectivity, access feasibility, route overlap, and numeric field accuracy.
\item \textbf{Validation.} We validate the dataset by training an LLM that achieves accurate map-free route generation, exhibits implicit spatial grounding from GPS coordinates to transit stations, and generalizes across diverse planning objectives with a single jointly trained model, confirming that the underlying transit knowledge is task-agnostic.
\end{itemize}

\section{Related Work}

\subsection{Transit Route Planning Methods}

Classical transit routing operates over explicit graph representations. Foundational algorithms such as Dijkstra \cite{dijkstra2022note} and A* \cite{hart1968formal} have been extended by transit-specific methods including RAPTOR \cite{delling2015round} and its Pareto-optimal extension \cite{delling2019fast}, Connection Scan Algorithm \cite{dibbelt2018csa}, and Transfer Patterns \cite{bast2010transfer}, enabling efficient multi-criteria journey planning on large-scale networks \cite{bast2016route}. All these approaches inherently require structured map infrastructure and real-time schedule data. Recent work explores whether LLMs can reduce this dependence. LLM-A* \cite{meng2024llmastar} incorporates LLM-generated heuristics into A* search but still requires the graph as input. GridRoute \cite{lin2025gridroute} benchmarks LLM path reasoning in synthetic grid environments. MapBench \cite{xing2025mapbench} and MapTrace \cite{chen2025maptrace} evaluate multimodal LLMs on pixel-level map navigation. ReasonMap \cite{feng2025can} targets transit map reading but reveals substantial limitations in visual reasoning accuracy. TraveLLM \cite{fang2025travellm} applies LLMs to transit disruption scenarios while remaining dependent on external map data. GenMRP \cite{wang2026genmrp} generates multiple routes within a dynamically constructed road sub-network and therefore still relies on an explicit road network. Across these efforts, no method has achieved end-to-end, map-free transit route generation from origin-destination information.

\subsection{Transit Data Sources}

Existing transit-related datasets each cover only partial aspects of the route planning problem. Vehicle trajectory datasets such as T-Drive \cite{yuan2010tdrive}, Porto Taxi \cite{moreira2013porto}, and GeoLife \cite{zheng2010geolife} record GPS traces of taxis or individuals \cite{zheng2015trajectory} but lack station structures, transfer logic, and line identifiers inherent to public transit. Static network datasets including GTFS \cite{wong2013leveraging}, OpenStreetMap \cite{haklay2008openstreetmap}, and CPTOND-2025 \cite{wang2026china} provide comprehensive topology and schedules across hundreds of cities but contain no user behavior or actual travel trajectories. No existing dataset combines complete route structures with behavioral annotations for data-driven transit route planning.

\subsection{Travel Planning and Routing Benchmarks}

Recent benchmarks evaluate LLM agents on planning and navigation tasks, yet none targets end-to-end transit route generation. TravelPlanner \cite{xie2024travelplanner}, NATURAL PLAN \cite{zheng2024naturalplan}, TripCraft \cite{chaudhuri2025tripcraft}, ChinaTravel \cite{shao2024chinatravel}, TripTailor \cite{wang2025triptailor}, TP-RAG \cite{chen2025tprag}, TravelBench \cite{cheng2025travelbench}, and TRIP-Bench \cite{shen2026tripbench} all focus on multi-day itinerary scheduling through tool-calling agents \cite{schick2023toolformer}, evaluating high-level constraint satisfaction rather than station-level route accuracy. Urban intelligence benchmarks such as CityBench \cite{feng2025citybench} and USTBench \cite{lai2025ustbench} cover diverse urban tasks but exclude or marginalize transit routing. MobilityBench \cite{song2026mobilitybench} is the closest to our setting, but it evaluates agent ability to orchestrate map APIs rather than to generate routes directly. No existing benchmark assesses whether an LLM can directly produce structurally valid transit routes with station-level precision.

\section{Dataset Construction}
\label{dataset}

\subsection{Data Collection}
\label{sec:collection}

TransitLM is constructed from public transit route planning logs provided by Amap, a leading navigation platform in China. We collect data from four major cities, Beijing, Shanghai, Shenzhen, and Chengdu, covering 120,845 stations and 13,666 bus and subway lines. From a single day of navigation logs we extract over 12.9 million planning sessions that form the pre-training corpus. Each session records origin and destination GPS coordinates, POI names, candidate routes with full station-ID sequences and line identifiers where stations are represented by unique numeric IDs rather than natural-language names, segment-level travel distances and times, route-type annotations, first/last-mile access details, and a user selection label identifying the route actually chosen for navigation. Since all candidate routes are generated by the production routing engine of the platform, they inherently satisfy connectivity and feasibility constraints, providing high-quality training signal without manual verification. All records are fully de-identified and privacy safeguards are detailed in Appendix~\ref{app:ethics}.

\textbf{Temporal split and de-duplication.} The single-day restriction applies only to the pre-training corpus and is a deliberate privacy-minimization choice, since multi-day continuous logs would make it far easier to reconstruct individual mobility routines such as home and work locations. The benchmark data are collected from later days, so every evaluation reported in this paper is a future-day setting rather than same-day memorization. We additionally remove from the official evaluation set any query whose directed origin--destination POI pair also appears in the pre-training corpus. Such repetition is rare to begin with, as directed POI-level OD sets drawn from adjacent days share only 4.97\% of their pairs on average over one month. The evaluation set is therefore disjoint from the pre-training corpus along both the temporal dimension and the OD dimension, so the reported accuracy cannot be explained by queries already seen during pre-training.

\begin{figure}[t]
\centering
\includegraphics[width=\textwidth]{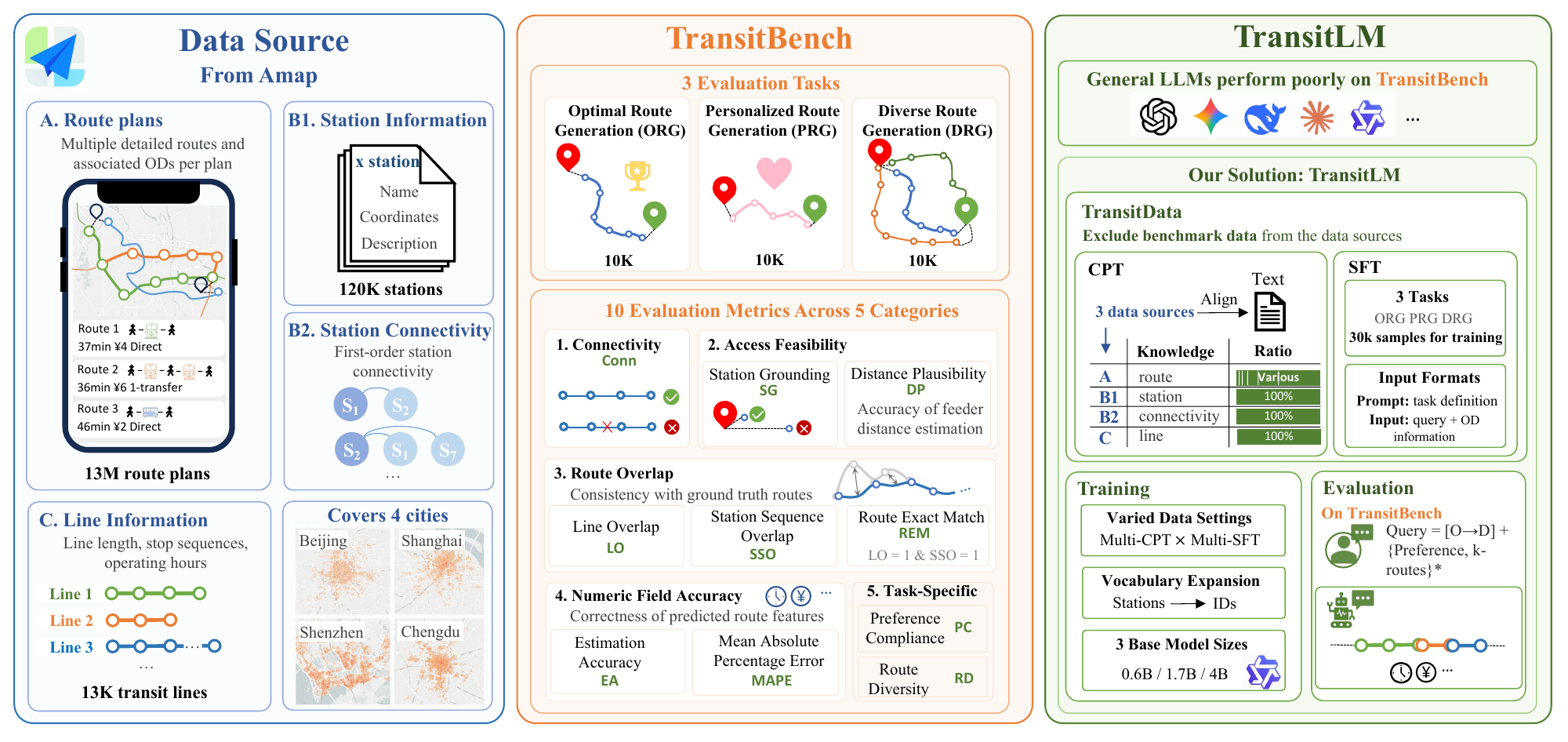}
\caption{Overview of TransitLM. \textbf{Left:} Data sources from Amap comprising route plans, station information, station connectivity, and line information across four cities. \textbf{Center:} TransitBench defines three evaluation tasks (ORG, PRG, DRG) with 10K test samples each, assessed by 10 metrics across five categories. \textbf{Right:} TransitLM addresses the limitations of general LLMs through continual pre-training on three knowledge sources and supervised fine-tuning on three tasks, with vocabulary expansion and varied data settings.}
\label{fig:method}
\end{figure}

\subsection{Data Schema}
\label{sec:schema}

\textbf{Continual Pre-Training (CPT) Corpus.} A textual corpus of 13.9 million records, comprising 12.9 million route planning sessions and 1.0 million static descriptions of stations and lines. Domain-adaptive continual pre-training \cite{gururangan2020dont} has proven effective for specializing language models to new domains. Each session record encodes a planning query as natural language: a query header specifying city, origin--destination coordinates, and POI names, followed by candidate routes with per-segment details. The user-selected route is placed first among the candidates, allowing the model to implicitly learn user preference patterns through next-token prediction. Static records describe individual lines and stations with attributes such as line length, stop sequences, operating hours, and connectivity. Representative examples of these record types are provided in Appendix~\ref{app:cpt_sample}. This formulation enables the model to internalize transit network topology and spatial relationships.

\textbf{Benchmark Supervised Fine-Tuning (SFT) Data.} Task-specific data constructed for three benchmark tasks (Section~\ref{benchmark}): optimal route generation, preference-aware planning, and multi-route generation. Each task selects specific routes from the candidate set according to task-defined criteria to construct structured labels. Each task provides 30,000 training and 10,000 test examples with task-specific filtering criteria. All examples follow a standardized prompt--label format as illustrated in Appendix~\ref{app:benchmark_examples}, enabling reproducible comparison across models and training configurations.

\subsection{Data Statistics and Analysis}
\label{sec:stats}

The CPT corpus comprises 13.9 million records from three complementary sources: 12,945,264 route planning sessions, 880,854 station descriptions, and 147,918 line descriptions. Table~\ref{tab:city_stats} summarizes key statistics across the four cities. Each session contains on average 6.32 candidate routes from the navigation engine; during CPT corpus construction, we retain at most five routes per session after diversity filtering.

\textbf{Route modality distribution.}
We classify each candidate route into four categories based on its transit segments, excluding walking which serves only as a connection between segments. Bus-only routes account for 33.0\%, subway-only for 19.0\%, and bus+subway for 16.8\%. Mixed routes, where at least one segment involves taxi or cycling as a first/last-mile connection to a transit line, represent 30.5\%. The remaining 0.7\% consist of non-transit alternatives such as taxi-only, or cycling-only routes. No single modality dominates the corpus, confirming balanced coverage across transit types.

\textbf{Route distance and travel time.}
Route distances span from under 5\,km to over 30\,km. Short-range routes within 5\,km account for 22.8\%, mid-range routes of 5--20\,km collectively represent 47.4\%, and long-range routes beyond 20\,km make up 29.7\%. Travel times exhibit a comparable spread, with the majority falling between 15 and 90 minutes. This breadth ensures that models trained on the corpus encounter the full continuum of urban commuting scenarios.

\textbf{Corpus sequence length.}
CPT records average 2,377 Chinese characters in length, with 58.4\% falling in the 2,000--5,000 range. Another 23.6\% lies between 1,000 and 2,000, while 2.4\% exceeds 5,000 characters, typically corresponding to long-distance routes with many intermediate stops. The corpus totals over 20 billion tokens, providing substantial training signal for continual pre-training~\cite{hoffmann2022training}.

\begin{table}[t]
\centering
\caption{CPT corpus statistics by city. Stations and Lines denote the number of unique entities covered. Routes/Sess.\ is the average number of candidate routes per session. Stops indicates the average station sequence length per route. Transfers, Fare are per-route averages.}
\label{tab:city_stats}
\small{%
\begin{tabular}{l c c c c c c c}
\toprule
\textbf{City} & \textbf{Sessions} & \textbf{Stations} & \textbf{Lines} & \textbf{Routes/Sess.} & \textbf{Stops} & \textbf{Transfers} & \textbf{Fare (\textyen)} \\
\midrule
Beijing   & 4,802,588 & 38,792 & 4,234 & 6.14 & 14.29 & 1.57 & 8.36 \\
Shanghai  & 3,144,747 & 35,617 & 3,702 & 6.54 & 13.42 & 1.51 & 8.38 \\
Shenzhen  & 2,877,449 & 14,980 & 2,307 & 6.41 & 12.45 & 1.23 & 7.17 \\
Chengdu   & 2,120,480 & 31,456 & 3,423 & 6.28 & 13.82 & 1.49 & 6.99 \\
\midrule
\rowcolor{gray!15}
Total & 12,945,264 & 120,845 & 13,666 & 6.32 & 13.58 & 1.47 & 7.88 \\
\bottomrule
\end{tabular}%
}
\end{table}

\section{Benchmark Tasks}
\label{benchmark}

End-to-end, map-free transit route planning requires a model to produce a complete route from a user query and origin-destination information alone, without relying on map infrastructure or routing engines. A complete route encompasses transit lines and station-ID sequences with transfer markers, from which the full trajectory can be reconstructed on a map for visualization, together with estimated distance, time, fare, and first/last-mile access details connecting the origin and destination to the transit network. These fields provide the essential information for a self-contained travel plan, covering both route reconstruction and travel decisions. To evaluate this capability under a standardized protocol, we design three benchmark tasks that collectively assess route accuracy, preference-conditioned planning, and output diversity.
\subsection{Task Definitions}
\begingroup
\setlength{\parskip}{0pt}
\paragraph{Optimal Route Generation.}
Given origin-destination information and a natural-language query, the model generates a single optimal transit route as structured JSON, including line sequence, station-ID sequence with transfer markers, distance, time, fare, and first/last-mile access details. The ground-truth label is the route the user selected and navigated, restricted to sessions where it is also the top-ranked option returned by the platform, so every label reflects a real traveler choice of verified quality.

\paragraph{Preference-Aware Planning.}
The input and output formats are identical to Optimal Route Generation, except that the query explicitly states a user preference. We define four preference categories that reflect the most common real-world planning needs: \textit{subway-first}, \textit{bus-first}, \textit{fewer transfers}, and \textit{shortest time}. The model must parse the stated preference from the query and generate a route that satisfies the corresponding constraint while remaining optimal under that criterion. Training data are constructed from sessions where the user explicitly set one of these preferences, and the ground-truth label follows the same two constraints as Optimal Route Generation.

\paragraph{Multi-Route Generation.}
Given the same OD input and a natural-language query, the model generates three diverse transit routes in a single JSON response. Each route shares the schema of Optimal Route Generation, with an additional \texttt{route\_tag} indicating the route type, formed by a primary mode label and an optional secondary access label. Ground-truth triples are assembled from the session's candidate pool by priority: (1)~the user-clicked route; (2)~routes with distinct tags or non-overlapping lines for diversity, selected in display order as ranked by the platform; and (3)~top-scored routes by an expert scoring function as fallback.
\endgroup

\begin{table}[t]
\centering
\caption{Comparison with general-purpose LLMs on \textbf{Optimal Route Generation} over 1{,}000 test samples across four cities. Column headers abbreviate full model names: GPT-5.4-pro, DeepSeek-V4-Pro, Gemini-3.1-Pro, Claude-Opus-4.6, Qwen3.6-Plus, and Doubao-Seed-2.0-Pro. $\uparrow$\,/\,$\downarrow$ indicate higher/lower is better. \textbf{Bold}: best; \underline{underline}: second best.}
\label{tab:llm_comparison}
\small
\setlength{\tabcolsep}{3pt}
\renewcommand{\arraystretch}{1.05}
\scalebox{1.05}{
\begin{tabular}{@{}l@{\enspace}c cccccc@{}}
\toprule
\textbf{Metric} & & \textbf{GPT-5.4} & \textbf{Deepseek-V4} & \textbf{Gemini-3.1} & \textbf{Claude-4.6} & \textbf{Qwen-3.6} & \textbf{Doubao} \\
\midrule
Connectivity & $\uparrow$ & 60.5\% & \underline{64.9\%} & \textbf{75.5\%} & 48.1\% & 45.0\% & 61.6\% \\
\cdashline{1-8}[0.4pt/3pt]
Station Grounding & $\uparrow$ & 60.5\% & 72.0\% & \textbf{93.9\%} & 50.7\% & 64.2\% & \underline{83.9\%} \\
Distance Plausibility & $\uparrow$ & 40.7\% & 49.9\% & \textbf{67.0\%} & 27.9\% & 42.9\% & \underline{59.4\%} \\
\cdashline{1-8}[0.4pt/3pt]
Line Overlap & $\uparrow$ & 0.407 & 0.447 & \textbf{0.600} & 0.396 & 0.475 & \underline{0.584} \\
Station Sequence Overlap & $\uparrow$ & 0.361 & 0.425 & \textbf{0.569} & 0.314 & 0.386 & \underline{0.517} \\
Route Exact Match & $\uparrow$ & 18.4\% & 23.7\% & \textbf{40.2\%} & 15.4\% & 19.8\% & \underline{33.6\%} \\
\cdashline{1-8}[0.4pt/3pt]
Estimation Accuracy & $\uparrow$ & 27.0\% & 30.5\% & \underline{38.9\%} & \textbf{47.3\%} & 20.2\% & 26.1\% \\
MAPE & $\downarrow$ & 13.7\% & 14.3\% & \textbf{10.7\%} & 12.1\% & 16.1\% & \underline{12.0\%} \\
\bottomrule
\end{tabular}
}
\end{table}

\subsection{Evaluation Metrics}
\label{sec:metrics}
\begingroup
\setlength{\parskip}{0pt}

We evaluate predicted routes along four complementary dimensions, supplemented by task-specific metrics. Formal definitions are provided in \textbf{Appendix~\ref{app:metrics}}.
\paragraph{Connectivity.} Verifies that every consecutive station pair in the predicted sequence is reachable via a shared transit line or a valid transfer. All subsequent metrics except task-specific ones are computed only on connected samples.
\paragraph{Access Feasibility.} Validates the first/last-mile segments connecting the origin/destination to the transit network. It comprises two sub-metrics: \textit{Station Grounding} (SG) checks whether the predicted boarding/alighting station is within a mode-specific distance threshold of the origin/destination, namely 3\,km for walking, 5\,km for cycling, and 10\,km for taxi, reflecting implicit spatial grounding \cite{li2023geolm} learned from training data; \textit{Distance Plausibility} (DP) verifies that the predicted access distance is physically plausible.
\paragraph{Route Overlap.} Quantifies the structural match between predicted and ground-truth routes using Intersection-over-Union (IoU). \textit{Line Overlap} (LO) computes IoU over the full line set including first/last-mile access segments; \textit{Station Sequence Overlap} (SSO) computes IoU over station ID sets; \textit{Route Exact Match} (REM) reports the fraction of samples achieving both LO\,=\,1 and SSO\,=\,1.
\paragraph{Numeric Field Accuracy.} Measures how accurately the model predicts route-level numeric attributes. Let $\mathcal{F} = \{\text{distance}, \text{time}, \text{fare}\}$ denote the set of numeric fields. \textit{Estimation Accuracy} (EA) measures the pass rate under a dual-tolerance criterion, and \textit{Mean Absolute Percentage Error} (MAPE) quantifies continuous error magnitude. Both are restricted to samples achieving REM (LO\,=\,1 and SSO\,=\,1), since the reference distance, time, and fare provide valid ground truth only for matching routes. Non-matching routes lack corresponding numeric ground truth for evaluation and are analyzed separately in Appendix~\ref{app:nonexact}.
\paragraph{Task-specific Metrics.} Preference-Aware Planning additionally uses \textit{Preference Compliance} (PC), which checks whether the predicted route satisfies the stated preference via hard rules. Multi-Route Generation uses \textit{Route Diversity} (RD), measuring the average pairwise line-set dissimilarity among the three generated routes; RD should be interpreted jointly with the four evaluation dimensions to balance diversity against route quality.
\endgroup

\section{Experiments}
\label{sec:experiments}

\subsection{Experimental Setup}
\label{sec:setup}

We use Qwen3-0.6B-Base, Qwen3-1.7B-Base, and Qwen3-4B-Base~\cite{yang2025qwen3} as backbones. We extend the vocabulary by registering all 120,845 station IDs as dedicated tokens, so that each station is represented as a single token. This prevents the model from hallucinating non-existent stations through character-level composition and enables it to learn station-level spatial and topological relationships directly. We do not explore larger models, as the 4B model already achieves strong performance across all tasks while larger variants would incur substantially higher training cost with diminishing returns.

Each model is trained through a two-stage pipeline. In the continual pre-training (CPT) stage \cite{gururangan2020dont}, all sequences are packed to a fixed length and optimized with cosine learning rate scheduling. In the subsequent supervised fine-tuning (SFT) stage \cite{ouyang2022training,wei2022finetuned}, each model is fine-tuned for one epoch on each benchmark task. The SFT data are drawn from a separate time period with no overlap with the CPT corpus, preventing data leakage. We additionally train a joint variant (Qwen3-4B-Joint) that fine-tunes the 4B CPT checkpoint on the combined SFT data of all three tasks, evaluating whether the transit knowledge learned during pre-training transfers across planning objectives, enabling unified deployment with a single model. All training is conducted on Alibaba Cloud PPU accelerators. Detailed hyperparameters are provided in Appendix~\ref{app:hyperparams}.

\subsection{Benchmark Results}
\label{sec:results}
\begingroup
\setlength{\parskip}{0pt}

\paragraph{Comparison with general-purpose LLMs.}
A central question underlying this dataset is whether existing general-purpose LLMs can perform transit route planning without domain-specific training data. We evaluate six state-of-the-art models on Optimal Route Generation over 1{,}000 test samples across four cities, as shown in Table~\ref{tab:llm_comparison}. To provide a maximally favorable setting, we simplify the output requirement: each model predicts only the boarding and alighting stations per leg, whereas our domain-specific models must generate the complete intermediate station sequence. This design isolates the core challenge of transit network knowledge from sequence-level generation difficulty, constituting a strictly more lenient evaluation. Despite this advantage, all models struggle substantially. The best performer, Gemini-3.1-Pro, achieves only 75.5\% connectivity and 40.2\% Route Exact Match, confirming that general-purpose LLMs lack the transit-specific topological knowledge for structurally valid route generation. The bottleneck lies in domain knowledge rather than model capacity or output complexity, underscoring the necessity of dedicated transit planning data.

\begin{table}[t]
\centering
\caption{Results on \textbf{Optimal Route Generation} with 10,000 test samples across four cities. Qwen3-4B-25 denotes CPT on 25\% of session data. Qwen3-4B-Joint is fine-tuned on the combined SFT data of all three tasks. $\uparrow$\,/\,$\downarrow$ indicate higher/lower is better. \textbf{Bold}: best; \underline{underline}: second best.}
\label{tab:task1_results}
\small
\setlength{\tabcolsep}{3pt}
\renewcommand{\arraystretch}{1.05}
\scalebox{1.05}{
\begin{tabular}{@{}l@{\enspace}c ccccc@{}}
\toprule
\textbf{Metric} & & \textbf{Qwen3-0.6B} & \textbf{Qwen3-1.7B} & \textbf{Qwen3-4B-25} & \textbf{Qwen3-4B} & \textbf{4B-Joint} \\
\midrule
Connectivity & $\uparrow$ & 93.5\% & 95.0\% & 95.9\% & \underline{97.0\%} & \textbf{97.9\%} \\
\cdashline{1-7}[0.4pt/3pt]
Station Grounding & $\uparrow$ & 96.3\% & 96.6\% & 97.7\% & \underline{98.5\%} & \textbf{98.9\%} \\
Distance Plausibility & $\uparrow$ & 84.9\% & 85.0\% & 87.4\% & \underline{91.0\%} & \textbf{92.9\%} \\
\cdashline{1-7}[0.4pt/3pt]
Line Overlap & $\uparrow$ & 0.812 & 0.827 & 0.811 & \underline{0.828} & \textbf{0.835} \\
Station Sequence Overlap & $\uparrow$ & 0.805 & 0.818 & 0.816 & \underline{0.838} & \textbf{0.847} \\
Route Exact Match & $\uparrow$ & 62.1\% & 64.1\% & 65.6\% & \underline{71.0\%} & \textbf{73.7\%} \\
\cdashline{1-7}[0.4pt/3pt]
Estimation Accuracy & $\uparrow$ & 97.2\% & 98.4\% & 97.8\% & \underline{98.5\%} & \textbf{98.6\%} \\
MAPE & $\downarrow$ & 2.03\% & 1.60\% & 1.88\% & \underline{1.33\%} & \textbf{1.30\%} \\
\bottomrule
\end{tabular}
}
\end{table}

\begin{table}[t]
\centering
\caption{Results on \textbf{Preference-Aware Planning} with 10,000 test samples across four cities. Qwen3-4B-25 denotes CPT on 25\% of session data. Qwen3-4B-Joint is fine-tuned on the combined SFT data of all three tasks. Label Preference Compliance is 96.02\%.}
\label{tab:task2_results}
\small
\setlength{\tabcolsep}{3pt}
\renewcommand{\arraystretch}{1.05}
\scalebox{1.05}{
\begin{tabular}{@{}l@{\enspace}c ccccc@{}}
\toprule
\textbf{Metric} & & \textbf{Qwen3-0.6B} & \textbf{Qwen3-1.7B} & \textbf{Qwen3-4B-25} & \textbf{Qwen3-4B} & \textbf{4B-Joint} \\
\midrule
Connectivity & $\uparrow$ & 85.0\% & 88.5\% & 90.1\% & \underline{93.2\%} & \textbf{95.3\%} \\
\cdashline{1-7}[0.4pt/3pt]
Station Grounding & $\uparrow$ & 92.2\% & 92.9\% & 94.2\% & \underline{96.5\%} & \textbf{97.4\%} \\
Distance Plausibility & $\uparrow$ & 73.3\% & 75.8\% & 77.4\% & \underline{84.6\%} & \textbf{88.5\%} \\
\cdashline{1-7}[0.4pt/3pt]
Line Overlap & $\uparrow$ & 0.650 & 0.656 & 0.683 & \underline{0.705} & \textbf{0.709} \\
Station Sequence Overlap & $\uparrow$ & 0.643 & 0.651 & 0.683 & \underline{0.716} & \textbf{0.726} \\
Route Exact Match & $\uparrow$ & 38.7\% & 39.4\% & 43.9\% & \underline{50.4\%} & \textbf{52.6\%} \\
\cdashline{1-7}[0.4pt/3pt]
Estimation Accuracy & $\uparrow$ & 90.8\% & 91.6\% & 90.9\% & \underline{92.5\%} & \textbf{92.8\%} \\
MAPE & $\downarrow$ & 2.74\% & 2.29\% & 2.49\% & \underline{2.05\%} & \textbf{1.88\%} \\
\cdashline{1-7}[0.4pt/3pt]
Preference Compliance & $\uparrow$ & 81.8\% & 81.6\% & 88.8\% & \underline{89.8\%} & \textbf{90.5\%} \\
\bottomrule
\end{tabular}
}
\end{table}

\paragraph{Main results.}
Tables~\ref{tab:task1_results}--\ref{tab:task3_results} report results on the three benchmark tasks. The Qwen3-4B model achieves $\geq$\,93\% connectivity, $\geq$\,96\% station grounding, and up to 71.0\% Route Exact Match, with estimation accuracy exceeding 92\% and MAPE below 2.1\% on exact-match samples. These results collectively confirm that end-to-end map-free route generation is feasible: the model not only produces connected routes but also grounds them to plausible stations, recovers correct complete routes at high rates, and accurately predicts numeric fields such as duration and walking distance. The high station grounding further suggests that implicit spatial grounding begins to emerge from training data, though the current evaluation includes origin and destination names alongside GPS coordinates. We provide stronger evidence for this capability in the GPS-only ablation below, where removing all textual cues yields minimal performance degradation for our models while general-purpose LLMs degrade substantially.

Station Sequence Overlap reaches 0.838 on Optimal Route Generation, 0.716 on Preference-Aware Planning, and 0.794 on Multi-Route Generation. The variation reflects task difficulty, as preference-conditioned planning must satisfy additional hard constraints such as minimum transfers or shortest time, while multi-route generation contends with higher label ambiguity due to multiple valid alternatives. Performance scales monotonically with model capacity, with the 4B model gaining +8.9pp Route Exact Match over the 0.6B on Optimal Route Generation. Even our smallest 0.6B model surpasses all six general-purpose LLMs evaluated under more lenient conditions as shown in Table~\ref{tab:llm_comparison}, underscoring that domain-specific data, rather than model scale, is the critical enabler.

The joint variant 4B-Joint further validates the generalizability of the learned transit knowledge. Trained on the combined SFT data of all three tasks, it matches or exceeds the single-task 4B counterpart on every metric across all benchmarks. The gains are most pronounced on Preference-Aware Planning, where connectivity improves by 2.1 percentage points and Route Exact Match by 2.2 percentage points, suggesting that exposure to diverse planning constraints strengthens the model's ability to satisfy individual task requirements. The complete absence of negative transfer on any metric confirms that the three tasks share underlying transit topology representations. Rather than competing for model capacity, the complementary planning objectives reinforce the shared spatial knowledge, confirming that the transit knowledge encoded in the dataset is task-agnostic and supports unified deployment with a single model.

\begin{table}[t]
\centering
\caption{Results on \textbf{Multi-Route Generation} with 10,000 test samples across four cities. Qwen3-4B-25 denotes CPT on 25\% of session data. Qwen3-4B-Joint is fine-tuned on the combined SFT data of all three tasks.}
\label{tab:task3_results}
\small
\setlength{\tabcolsep}{3pt}
\renewcommand{\arraystretch}{1.05}
\scalebox{1.05}{
\begin{tabular}{@{}l@{\enspace}c ccccc@{}}
\toprule
\textbf{Metric} & & \textbf{Qwen3-0.6B} & \textbf{Qwen3-1.7B} & \textbf{Qwen3-4B-25} & \textbf{Qwen3-4B} & \textbf{4B-Joint} \\
\midrule
Connectivity & $\uparrow$ & 92.3\% & 95.2\% & 94.9\% & \underline{96.3\%} & \textbf{97.1\%} \\
\cdashline{1-7}[0.4pt/3pt]
Station Grounding & $\uparrow$ & 95.4\% & 95.6\% & 97.0\% & \underline{98.0\%} & \textbf{98.8\%} \\
Distance Plausibility & $\uparrow$ & 81.0\% & 80.6\% & 84.8\% & \underline{90.1\%} & \textbf{91.7\%} \\
\cdashline{1-7}[0.4pt/3pt]
Line Overlap & $\uparrow$ & 0.753 & 0.757 & 0.756 & \underline{0.782} & \textbf{0.794} \\
Station Sequence Overlap & $\uparrow$ & 0.747 & 0.750 & 0.761 & \underline{0.794} & \textbf{0.808} \\
Route Exact Match & $\uparrow$ & 52.6\% & 52.9\% & 57.9\% & \underline{64.5\%} & \textbf{67.2\%} \\
\cdashline{1-7}[0.4pt/3pt]
Estimation Accuracy & $\uparrow$ & 96.3\% & 97.6\% & 97.4\% & \underline{98.0\%} & \textbf{98.1\%} \\
MAPE & $\downarrow$ & 2.28\% & 1.74\% & 1.90\% & \underline{1.45\%} & \textbf{1.33\%} \\
\cdashline{1-7}[0.4pt/3pt]
Route Diversity & $\uparrow$ & 0.514 & 0.532 & 0.534 & \underline{0.545} & \textbf{0.547} \\
\bottomrule
\end{tabular}
}
\end{table}

\begin{table}[t]
\centering
\caption{Data scaling on \textbf{Optimal Route Generation}: Qwen3-4B trained with varying CPT session data fractions. All variants share identical static descriptions and SFT data.}
\label{tab:scaling}
\small
\setlength{\tabcolsep}{3pt}
\renewcommand{\arraystretch}{1.05}
\scalebox{1.05}{
\begin{tabular}{@{}l@{\enspace}c ccccc@{}}
\toprule
\textbf{Metric} & & \textbf{4B-6.25\%} & \textbf{4B-12.5\%} & \textbf{4B-25\%} & \textbf{4B-50\%} & \textbf{4B-100\%} \\
\midrule
Connectivity & $\uparrow$ & 94.0\% & 95.4\% & 95.9\% & \underline{96.8\%} & \textbf{97.0\%} \\
\cdashline{1-7}[0.4pt/3pt]
Station Grounding & $\uparrow$ & 93.5\% & 96.4\% & 97.7\% & \underline{97.9\%} & \textbf{98.5\%} \\
Distance Plausibility & $\uparrow$ & 76.2\% & 84.7\% & 87.4\% & \underline{89.3\%} & \textbf{91.0\%} \\
\cdashline{1-7}[0.4pt/3pt]
Line Overlap & $\uparrow$ & 0.738 & 0.789 & 0.811 & \underline{0.825} & \textbf{0.828} \\
Station Sequence Overlap & $\uparrow$ & 0.731 & 0.790 & 0.816 & \underline{0.829} & \textbf{0.838} \\
Route Exact Match & $\uparrow$ & 49.9\% & 61.2\% & 65.6\% & \underline{68.9\%} & \textbf{71.0\%} \\
\cdashline{1-7}[0.4pt/3pt]
Estimation Accuracy & $\uparrow$ & 92.2\% & 96.6\% & 97.8\% & \underline{98.3\%} & \textbf{98.5\%} \\
MAPE & $\downarrow$ & 3.26\% & 2.27\% & 1.88\% & \underline{1.51\%} & \textbf{1.33\%} \\
\bottomrule
\end{tabular}
}
\end{table}

\paragraph{Data scaling.}
To examine how CPT data volume affects performance, we train Qwen3-4B on four reduced session data fractions (6.25\%, 12.5\%, 25\%, 50\%) with 100\% as the reference, while retaining all static descriptions and SFT data unchanged. Table~\ref{tab:scaling} reports results on Optimal Route Generation, and results on the remaining two tasks are provided in Appendix~\ref{app:scaling_additional}. All metrics improve monotonically with data volume, confirming that the current dataset scale is well-justified. Notably, even at 6.25\% of the CPT session data, the model already achieves 94.0\% connectivity and 49.9\% Route Exact Match, demonstrating that end-to-end map-free route planning is practically viable with modest data collection effort. Different metrics exhibit distinct data sensitivity, revealing a clear learning hierarchy: basic network topology is acquired first, with connectivity reaching 94\% at the smallest fraction, while precise route matching and numeric calibration are substantially more data-hungry, as Route Exact Match drops by 21.1 percentage points and MAPE increases from 1.33\% to 3.26\% at 6.25\%. This pattern suggests that the model learns the structural ``grammar'' of transit networks rapidly but requires denser coverage to master fine-grained route preferences and distance estimation. Consistent with this pattern, increasing the CPT session data fraction from 50\% to 100\% raises Route Exact Match by a further 2.1 percentage points but Connectivity by only 0.2 points.

\paragraph{GPS-only ablation.}
To disentangle the contribution of spatial knowledge acquired during training from that of textual cues present in the input query, we remove all natural-language queries and retain only origin--destination GPS coordinates as input. Tables~\ref{tab:gps_llm} and~\ref{tab:gps_ours} report results on Optimal Route Generation under this setting. General-purpose LLMs mostly drop to $<$\,1\% Route Exact Match, indicating their route planning relies on textual semantics of origin and destination names rather than spatial understanding of coordinates. Their connectivity actually \emph{increases} under GPS-only input, e.g., DeepSeek-V4 rises from 64.9\% to 80.3\%, yet Station Grounding plummets from 72.0\% to 16.8\%, confirming that without textual cues LLMs cannot ground the query spatially and fall back to memorized high-frequency stations. In contrast, our domain-specific models exhibit near-zero degradation. Qwen3-4B retains 70.4\% Route Exact Match compared to 71.0\% with text, and 4B-Joint retains 72.9\% compared to 73.7\%, demonstrating that the planning capability is grounded in spatial representations learned through CPT rather than dependent on textual input.

\begin{table}[t]
\centering
\caption{GPS-only ablation on general-purpose LLMs for \textbf{Optimal Route Generation} over 1{,}000 test samples across four cities. All textual cues are removed and only GPS coordinates are provided as input. Estimation Accuracy and MAPE are omitted because Route Exact Match samples are too few to yield reliable estimates. Column headers abbreviate full model names as in Table~\ref{tab:llm_comparison}.}
\label{tab:gps_llm}
\small
\setlength{\tabcolsep}{3pt}
\renewcommand{\arraystretch}{1.05}
\scalebox{1.05}{
\begin{tabular}{@{}l@{\enspace}c cccccc@{}}
\toprule
\textbf{Metric} & & \textbf{GPT-5.4} & \textbf{Deepseek-V4} & \textbf{Gemini-3.1} & \textbf{Claude-4.6} & \textbf{Qwen-3.6} & \textbf{Doubao} \\
\midrule
Connectivity & $\uparrow$ & 78.9\% & \underline{80.3\%} & \textbf{81.3\%} & 64.9\% & 65.5\% & 73.7\% \\
\cdashline{1-8}[0.4pt/3pt]
Station Grounding & $\uparrow$ & 14.4\% & 16.8\% & \textbf{79.9\%} & 15.3\% & 11.3\% & \underline{24.1\%} \\
Distance Plausibility & $\uparrow$ & 2.0\% & \underline{5.3\%} & \textbf{33.3\%} & 1.2\% & 0.5\% & 1.6\% \\
\cdashline{1-8}[0.4pt/3pt]
Line Overlap & $\uparrow$ & 0.197 & 0.191 & \textbf{0.457} & 0.170 & 0.179 & \underline{0.241} \\
Station Sequence Overlap & $\uparrow$ & 0.055 & 0.060 & \textbf{0.326} & 0.040 & 0.037 & \underline{0.066} \\
Route Exact Match & $\uparrow$ & \underline{0.6\%} & \underline{0.6\%} & \textbf{17.7\%} & \underline{0.6\%} & 0.2\% & 0.5\% \\
\bottomrule
\end{tabular}
}
\end{table}

\begin{table}[t]
\centering
\caption{GPS-only ablation on our domain-specific models for \textbf{Optimal Route Generation} with 10,000 test samples across four cities. Only raw GPS coordinates are provided as input.}
\label{tab:gps_ours}
\small
\setlength{\tabcolsep}{3pt}
\renewcommand{\arraystretch}{1.05}
\scalebox{1.05}{
\begin{tabular}{@{}l@{\enspace}c ccccc@{}}
\toprule
\textbf{Metric} & & \textbf{Qwen3-0.6B} & \textbf{Qwen3-1.7B} & \textbf{Qwen3-4B-25} & \textbf{Qwen3-4B} & \textbf{4B-Joint} \\
\midrule
Connectivity & $\uparrow$ & 93.5\% & 95.6\% & 95.8\% & \underline{97.2\%} & \textbf{98.0\%} \\
\cdashline{1-7}[0.4pt/3pt]
Station Grounding & $\uparrow$ & 95.9\% & 96.4\% & 97.7\% & \underline{98.3\%} & \textbf{98.8\%} \\
Distance Plausibility & $\uparrow$ & 83.6\% & 84.4\% & 87.0\% & \underline{90.4\%} & \textbf{92.6\%} \\
\cdashline{1-7}[0.4pt/3pt]
Line Overlap & $\uparrow$ & 0.805 & 0.814 & 0.807 & \underline{0.821} & \textbf{0.827} \\
Station Sequence Overlap & $\uparrow$ & 0.797 & 0.805 & 0.810 & \underline{0.832} & \textbf{0.839} \\
Route Exact Match & $\uparrow$ & 61.0\% & 62.5\% & 65.1\% & \underline{70.4\%} & \textbf{72.9\%} \\
\cdashline{1-7}[0.4pt/3pt]
Estimation Accuracy & $\uparrow$ & 95.9\% & 97.5\% & 96.8\% & \underline{97.6\%} & \textbf{98.1\%} \\
MAPE & $\downarrow$ & 2.44\% & 2.07\% & 2.24\% & \underline{1.75\%} & \textbf{1.52\%} \\
\bottomrule
\end{tabular}
}
\end{table}
\endgroup

\begin{table}[t]
\centering
\caption{Full station-sequence comparison on \textbf{Optimal Route Generation} with 10,000 test samples across four cities, with every model producing complete intermediate station sequences. Qwen3-4B and 4B-Joint are TransitLM variants. $\dagger$ marks tool-augmented models, which receive routing-engine candidates in context. \textbf{Bold}: best; \underline{underline}: second best.}
\label{tab:fair_sequence}
\small
\setlength{\tabcolsep}{3pt}
\renewcommand{\arraystretch}{1.05}
\scalebox{1.05}{
\begin{tabular}{@{}l@{\enspace}c ccccc@{}}
\toprule
\textbf{Metric} & & \textbf{GPT-5.4$^{\dagger}$} & \textbf{Claude-4.6$^{\dagger}$} & \textbf{Gemini-3.1$^{\dagger}$} & \textbf{Qwen3-4B} & \textbf{4B-Joint} \\
\midrule
Connectivity & $\uparrow$ & 97.2\% & \textbf{98.1\%} & 96.7\% & 97.0\% & \underline{97.9\%} \\
\cdashline{1-7}[0.4pt/3pt]
Station Grounding & $\uparrow$ & 98.5\% & \textbf{100.0\%} & \textbf{100.0\%} & 98.5\% & \underline{98.9\%} \\
Distance Plausibility & $\uparrow$ & 92.7\% & \underline{97.2\%} & \textbf{99.4\%} & 91.0\% & 92.9\% \\
\cdashline{1-7}[0.4pt/3pt]
Line Overlap & $\uparrow$ & 0.833 & 0.816 & \underline{0.834} & 0.828 & \textbf{0.835} \\
Station Sequence Overlap & $\uparrow$ & \underline{0.849} & 0.838 & \textbf{0.851} & 0.838 & 0.847 \\
Route Exact Match & $\uparrow$ & 72.3\% & 72.1\% & \underline{72.6\%} & 71.0\% & \textbf{73.7\%} \\
\bottomrule
\end{tabular}
}
\end{table}

\paragraph{Comparison with tool-augmented LLMs.}
General-purpose LLMs can access routing tools, making tool-augmented models a stronger baseline than standalone LLMs. We test whether transit knowledge learned from data can yield route quality comparable to these engine-assisted models without engine access at inference time. We provide GPT-5.4, Claude-4.6, and Gemini-3.1 with routing-engine candidates for reference, while TransitLM receives no candidates. All models must output complete intermediate station sequences, and Table~\ref{tab:fair_sequence} reports results on Optimal Route Generation. TransitLM-4B-Joint achieves 0.847 Station Sequence Overlap, only 0.004 below the best tool-augmented result, and 73.7\% Route Exact Match, 1.1 percentage points above the best tool-augmented result. The single-task Qwen3-4B also reaches 0.838 and 71.0\%, respectively. However, the best tool-augmented models achieve higher Station Grounding and Distance Plausibility, indicating trade-offs across metrics rather than a uniform advantage. These results provide further evidence that training on TransitLM supports end-to-end map-free route generation with quality comparable to tool-augmented LLMs on this benchmark, without external routing-engine access at inference time.

\paragraph{Additional experiments.} CPT training exhibits rapid early convergence and sustained loss reduction, with larger models attaining lower training loss, as shown in Appendix~\ref{app:cpt_loss}. Under a fixed CPT data budget, four-city training largely preserves Connectivity and Station Grounding despite a 3.1-fold expansion of the station vocabulary, with a reduction in Route Exact Match, as reported in Appendix~\ref{app:single_city}. The CPT ablation in Appendix~\ref{app:cpt_ablation} shows that SFT-only leads in standard-input Route Exact Match but suffers greater degradation under GPS-only input than CPT-based models, distinguishing accuracy with textual cues from robustness without them. Data-scaling gains and GPS-only robustness also extend to Preference-Aware Planning and Multi-Route Generation, as detailed in Appendix~\ref{app:scaling_additional}. Temporal tests in Appendix~\ref{app:temporal} show stable performance for Qwen3-4B without retraining three months later and across weekdays, weekends, and request times. Sparse-region tests and a Beijing SFT holdout with Beijing retained in CPT show larger declines in Route Exact Match than in Connectivity and Station Grounding, as examined in Appendix~\ref{app:geographic}. Appendix~\ref{app:rag_comparison} reports that tool access raises Route Exact Match for all six models and narrows gaps under identical test samples and simplified requirements as Table~\ref{tab:llm_comparison}. Finally, automated checks in Appendix~\ref{app:nonexact} show that connected, non-exact TransitLM outputs can remain spatially plausible and overlap closely with alternative routes for the same query, while limited exact candidate matching suggests generation beyond simple reproduction.

\section{Conclusion}
\label{sec:conclusion}

We presented TransitLM, a large-scale dataset of nearly 13 million transit route planning records across four Chinese cities, together with a three-task benchmark and evaluation metrics that establish a standardized protocol for map-free transit route generation. Our experiments demonstrate that end-to-end transit route planning through pure text generation is feasible without any external map or routing engine: the topological, spatial, and behavioral knowledge required can be acquired entirely from data. The resulting representations capture genuine spatial structure rather than depending on textual cues in the input query, as evidenced by near-zero performance degradation under GPS-only input where most general-purpose LLMs collapse. Joint training further confirms that the acquired knowledge is task-agnostic, as the three planning capabilities reinforce each other with no negative transfer. The current release covers four Chinese cities and represents a static snapshot of transit route structures. Future directions include extending geographic coverage and incorporating real-time dynamics, as discussed further in Appendix~\ref{app:limitations}.

\begin{ack}
This work was supported by Amap, Alibaba Group. No external grant funding was
received.

Competing interests: All authors are employees of Amap, Alibaba Group, which
operates the navigation platform underlying the TransitLM dataset. The authors
declare no other competing interests.
\end{ack}

{
\small
\bibliographystyle{plainnat}
\bibliography{references}

@article{openai2023gpt4,
  title={{GPT}-4 technical report},
  author={Achiam, Josh and Adler, Steven and Agarwal, Sandhini and Ahmad, Lama and Akkaya, Ilge and Aleman, Florencia Leoni and Almeida, Diogo and Altenschmidt, Janko and Altman, Sam and Anadkat, Shyamal and others},
  journal={arXiv preprint arXiv:2303.08774},
  year={2023}
}

@article{yang2025qwen3,
  title={{Qwen3} technical report},
  author={Yang, An and Li, Anfeng and Yang, Baosong and Zhang, Beichen and Hui, Binyuan and Zheng, Bo and Yu, Bowen and Gao, Chang and Huang, Chengen and Lv, Chenxu and others},
  journal={arXiv preprint arXiv:2505.09388},
  year={2025}
}

@inproceedings{yuan2010tdrive,
  title={{T}-Drive: Driving directions based on taxi trajectories},
  author={Yuan, Jing and Zheng, Yu and Zhang, Chengyang and Xie, Wenlei and Xie, Xing and Sun, Guangzhong and Huang, Yan},
  booktitle={Proceedings of the 18th ACM SIGSPATIAL International Conference on Advances in Geographic Information Systems},
  pages={99--108},
  year={2010}
}

@article{moreira2013porto,
  title={Predicting taxi--passenger demand using streaming data},
  author={Moreira-Matias, Luis and Gama, Joao and Ferreira, Michel and Mendes-Moreira, Joao and Damas, Luis},
  journal={IEEE Transactions on Intelligent Transportation Systems},
  volume={14},
  number={3},
  pages={1393--1402},
  year={2013}
}

@article{wong2013leveraging,
  title={Leveraging the general transit feed specification for efficient transit analysis},
  author={Wong, James},
  journal={Transportation Research Record},
  volume={2338},
  number={1},
  pages={11--19},
  year={2013}
}

@article{wang2026china,
  title={China Public Transport Operation Network Dataset ({CPTOND}-2025): National-scale bus-metro vector dataset},
  author={Wang, Liang and Wei, He and Guan, Yu and Ouyang, Libin and Xu, DanDan and Han, Xuehua and Zhang, Min and Chen, Meng and Sun, Daosheng and Gong, Daqing and others},
  journal={Scientific Data},
  year={2026}
}

@incollection{dijkstra2022note,
  title={A note on two problems in connexion with graphs},
  author={Dijkstra, Edsger W},
  booktitle={Edsger Wybe Dijkstra: his life, work, and legacy},
  pages={287--290},
  year={2022}
}

@article{hart1968formal,
  title={A formal basis for the heuristic determination of minimum cost paths},
  author={Hart, Peter E and Nilsson, Nils J and Raphael, Bertram},
  journal={IEEE Transactions on Systems Science and Cybernetics},
  volume={4},
  number={2},
  pages={100--107},
  year={1968}
}

@inproceedings{fang2025travellm,
  title={{TraveLLM}: Could you plan my public transit alternatives in face of a network disruption?},
  author={Fang, Bowen and Yang, Zixiao and Di, Xuan},
  booktitle={2025 IEEE 28th International Conference on Intelligent Transportation Systems (ITSC)},
  pages={4711--4717},
  year={2025}
}

@article{feng2025can,
  title={Can {MLLMs} guide me home? {A} benchmark study on fine-grained visual reasoning from transit maps},
  author={Feng, Sicheng and Wang, Song and Ouyang, Shuyi and Kong, Lingdong and Song, Zikai and Zhu, Jianke and Wang, Huan and Wang, Xinchao},
  journal={arXiv preprint arXiv:2505.18675},
  year={2025}
}

@inproceedings{feng2025citybench,
  title={{CityBench}: Evaluating the capabilities of large language models for urban tasks},
  author={Feng, Jie and Zhang, Jun and Liu, Tianhui and Zhang, Xin and Ouyang, Tianjian and Yan, Junbo and Du, Yuwei and Guo, Siqi and Li, Yong},
  booktitle={Proceedings of the 31st ACM SIGKDD Conference on Knowledge Discovery and Data Mining V. 2},
  pages={5413--5424},
  year={2025}
}

@article{zheng2015trajectory,
  title={Trajectory data mining: an overview},
  author={Zheng, Yu},
  journal={ACM Transactions on Intelligent Systems and Technology},
  volume={6},
  number={3},
  pages={1--41},
  year={2015}
}

@article{zheng2010geolife,
  title={{GeoLife}: A collaborative social networking service among user, location and trajectory},
  author={Zheng, Yu and Xie, Xing and Ma, Wei-Ying},
  journal={IEEE Data Engineering Bulletin},
  volume={33},
  number={2},
  pages={32--39},
  year={2010}
}

@article{haklay2008openstreetmap,
  title={{OpenStreetMap}: User-generated street maps},
  author={Haklay, Mordechai and Weber, Patrick},
  journal={IEEE Pervasive Computing},
  volume={7},
  number={4},
  pages={12--18},
  year={2008}
}

@article{delling2015round,
  title={Round-based public transit routing},
  author={Delling, Daniel and Pajor, Thomas and Werneck, Renato F},
  journal={Transportation Science},
  volume={49},
  number={3},
  pages={591--604},
  year={2015}
}

@article{dibbelt2018csa,
  title={Connection scan algorithm},
  author={Dibbelt, Julian and Pajor, Thomas and Strasser, Ben and Wagner, Dorothea},
  journal={Journal of Experimental Algorithmics},
  volume={23},
  pages={1--56},
  year={2018}
}

@inproceedings{bast2010transfer,
  title={Fast routing in very large public transportation networks using transfer patterns},
  author={Bast, Hannah and Carlsson, Erik and Eigenwillig, Arno and Geisberger, Robert and Harrelson, Chris and Raychev, Veselin and Viger, Fabien},
  booktitle={European Symposium on Algorithms},
  pages={290--301},
  year={2010}
}

@incollection{bast2016route,
  title={Route planning in transportation networks},
  author={Bast, Hannah and Delling, Daniel and Goldberg, Andrew and M{\"u}ller-Hannemann, Matthias and Pajor, Thomas and Sanders, Peter and Wagner, Dorothea and Werneck, Renato F},
  booktitle={Algorithm engineering: Selected results and surveys},
  pages={19--80},
  year={2016}
}

@inproceedings{guo2026towards,
  title={Towards Full Candidate Interaction: A Comprehensive Comparison Network for Better Route Recommendation},
  author={Guo, Hanyu and Chen, Chao and Xu, Longfei and Wang, Chengzhang and Liu, Kaikui and Chu, Xiangxiang},
  booktitle={Proceedings of the 20th ACM Conference on Recommender Systems},
  pages={1550--1558},
  year={2026}
}

@inproceedings{wang2026genmrp,
  title={{GenMRP}: A Generative Multi-Route Planning Framework for Efficient and Personalized Real-Time Industrial Navigation},
  author={Wang, Chengzhang and Chen, Chao and Tao, Jun and Liu, Tengfei and Bai, He and Wang, Song and Xu, Longfei and Liu, Kaikui and Chu, Xiangxiang},
  booktitle={Proceedings of the 35th ACM International Conference on Information and Knowledge Management},
  pages={8579--8586},
  year={2026}
}

@article{chen2026scasrec,
  title={{SCASRec}: A Self-Correcting and Auto-Stopping Model for Generative Route List Recommendation},
  author={Chen, Chao and Xu, Longfei and Su, Daohan and Liu, Tengfei and Guo, Hanyu and Duan, Yihai and Liu, Kaikui and Chu, Xiangxiang},
  journal={arXiv preprint arXiv:2602.03324},
  year={2026}
}

@inproceedings{yu2025dsfnet,
  title={{DSFNet}: Learning Disentangled Scenario Factorization for Multi-Scenario Route Ranking},
  author={Yu, Jiahao and Duan, Yihai and Xu, Longfei and Chen, Chao and Liu, Shuliang and Liu, Kaikui and Yang, Fan and Chu, Xiangxiang and Guo, Ning},
  booktitle={Companion Proceedings of the ACM Web Conference 2025},
  pages={567--576},
  year={2025}
}

@inproceedings{meng2024llmastar,
  title={{LLM-A}*: Large language model enhanced incremental heuristic search on path planning},
  author={Meng, Silin and Wang, Yiwei and Yang, Cheng-Fu and Peng, Nanyun and Chang, Kai-Wei},
  booktitle={Findings of the Association for Computational Linguistics: EMNLP 2024},
  pages={1087--1102},
  year={2024}
}

@article{lin2025gridroute,
  title={{GridRoute}: A benchmark for {LLM}-based route planning with cardinal movement in grid environments},
  author={Li, Kechen and Tao, Yaotian and Wen, Ximing and Sun, Quanwei and Gong, Zifei and Xu, Chang and Zhang, Xizhe and Ji, Tianbo},
  journal={arXiv preprint arXiv:2505.24306},
  year={2025}
}

@article{chen2025maptrace,
  title={{MapTrace}: Scalable data generation for route tracing on maps},
  author={Panagopoulou, Artemis and Purohit, Aveek and Kulshrestha, Achin and Yazdani, Soroosh and Goyal, Mohit},
  journal={arXiv preprint arXiv:2512.19609},
  year={2025}
}

@article{xing2025mapbench,
  title={{MapBench}: Can large vision language models read maps like a human?},
  author={Xing, Shuo and Sun, Zezhou and Xie, Shuangyu and Chen, Kaiyuan and Huang, Yanjia and Wang, Yuping and Li, Jiachen and Song, Dezhen and Tu, Zhengzhong},
  journal={arXiv preprint arXiv:2503.14607},
  year={2025}
}

@inproceedings{xie2024travelplanner,
  title={{TravelPlanner}: A benchmark for real-world planning with language agents},
  author={Xie, Jian and Zhang, Kai and Chen, Jiangjie and Zhu, Tinghui and Lou, Renze and Tian, Yuandong and Xiao, Yanghua and Su, Yu},
  booktitle={International Conference on Machine Learning},
  pages={54590--54613},
  year={2024}
}

@article{zheng2024naturalplan,
  title={{NATURAL PLAN}: Benchmarking {LLMs} on natural language planning},
  author={Zheng, Huaixiu Steven and Mishra, Swaroop and Zhang, Hugh and Chen, Xinyun and Chen, Minmin and Nova, Azade and Hou, Le and Cheng, Heng-Tze and Le, Quoc V and Chi, Ed H and others},
  journal={arXiv preprint arXiv:2406.04520},
  year={2024}
}

@inproceedings{chaudhuri2025tripcraft,
  title={{TripCraft}: A benchmark for spatio-temporally fine grained travel planning},
  author={Chaudhuri, Soumyabrata and Purkar, Pranav and Raghav, Ritwik and Mallick, Shubhojit and Gupta, Manish and Jana, Abhik and Ghosh, Shreya},
  booktitle={Proceedings of the 63rd Annual Meeting of the Association for Computational Linguistics},
  pages={17035--17064},
  year={2025}
}

@inproceedings{shao2024chinatravel,
  title={{ChinaTravel}: An open-ended benchmark for language agents in {Chinese} travel planning},
  author={Shao, Jie-Jing and Zhang, Bo-Wen and Yang, Xiao-Wen and Chen, Baizhi and Han, Siyu and Wei, Wen-Da and Cai, Guohao and Dong, Zhenhua and Guo, Lan-Zhe and Li, Yu-Feng},
  booktitle={NeurIPS 2025 Workshop on Evaluating the Evolving LLM Lifecycle},
  year={2025}
}

@inproceedings{wang2025triptailor,
  title={{TripTailor}: A real-world benchmark for personalized travel planning},
  author={Wang, Kaimin and Shen, Yuanzhe and Lv, Changze and Zheng, Xiaoqing and Huang, Xuan-Jing},
  booktitle={Findings of the Association for Computational Linguistics: ACL 2025},
  pages={9705--9723},
  year={2025}
}

@inproceedings{chen2025tprag,
  title={{TP-RAG}: Benchmarking retrieval-augmented large language model agents for spatiotemporal-aware travel planning},
  author={Ni, Hang and Liu, Fan and Ma, Xinyu and Su, Lixin and Wang, Shuaiqiang and Yin, Dawei and Xiong, Hui and Liu, Hao},
  booktitle={Proceedings of the 2025 Conference on Empirical Methods in Natural Language Processing},
  pages={12403--12429},
  year={2025}
}

@article{lai2025ustbench,
  title={{USTBench}: Benchmarking and dissecting spatiotemporal reasoning of {LLMs} as urban agents},
  author={Lai, Siqi and Ning, Yansong and Yuan, Zirui and Chen, Zhixi and Liu, Hao},
  journal={arXiv preprint arXiv:2505.17572},
  year={2025}
}

@article{shen2026tripbench,
  title={{TRIP-Bench}: A benchmark for long-horizon interactive agents in real-world scenarios},
  author={Shen, Yuanzhe and Huang, Zisu and Wang, Zhengyuan and others},
  journal={arXiv preprint arXiv:2602.01675},
  year={2026}
}

@article{cheng2025travelbench,
  title={{TravelBench}: A real-world benchmark for multi-turn and tool-augmented travel planning},
  author={Cheng, Xiang and Hu, Yulan and Zhang, Xiangwen and Xu, Lu and Pan, Zheng and Li, Xin and Liu, Yong},
  journal={arXiv preprint arXiv:2512.22673},
  year={2025}
}

@article{song2026mobilitybench,
  title={{MobilityBench}: A benchmark for evaluating route-planning agents in real-world mobility scenarios},
  author={Song, Zhiheng and Zhang, Jingshuai and Qin, Chuan and others},
  journal={arXiv preprint arXiv:2602.22638},
  year={2026}
}

@article{yan2026inttravel,
  title={{IntTravel}: A real-world dataset and generative framework for integrated multi-task travel recommendation},
  author={Yan, Huimin and Xu, Longfei and Sun, Junjie and Liu, Zheng and Luo, Wei and Liu, Kaikui and Chu, Xiangxiang},
  journal={arXiv preprint arXiv:2602.11664},
  year={2026}
}

@inproceedings{gururangan2020dont,
  title={Don’t stop pretraining: Adapt language models to domains and tasks},
  author={Gururangan, Suchin and Marasovi{\'c}, Ana and Swayamdipta, Swabha and Lo, Kyle and Beltagy, Iz and Downey, Doug and Smith, Noah A},
  booktitle={Proceedings of the 58th Annual Meeting of the Association for Computational Linguistics},
  pages={8342--8360},
  year={2020}
}

@inproceedings{ouyang2022training,
  title={Training language models to follow instructions with human feedback},
  author={Ouyang, Long and Wu, Jeffrey and Jiang, Xu and others},
  booktitle={Advances in Neural Information Processing Systems},
  volume={35},
  pages={27730--27744},
  year={2022}
}

@inproceedings{wei2022finetuned,
  title={Finetuned language models are zero-shot learners},
  author={Wei, Jason and Bosma, Maarten and Zhao, Vincent and Guu, Kelvin and Yu, Adams Wei and Lester, Brian and Du, Nan and Dai, Andrew M and Le, Quoc V},
  booktitle={International Conference on Learning Representations},
  year={2022}
}

@inproceedings{kambhampati2024llms,
  title={Position: {LLMs} can't plan, but can help planning in {LLM}-modulo frameworks},
  author={Kambhampati, Subbarao and Valmeekam, Karthik and Guan, Lin and Verma, Mudit and Stechly, Kaya and Bhambri, Siddhant and Saldyt, Lucas Paul and Murthy, Anil B},
  booktitle={International Conference on Machine Learning},
  year={2024}
}

@inproceedings{valmeekam2023planning,
  title={On the planning abilities of large language models -- a critical investigation},
  author={Valmeekam, Karthik and Marquez, Matthew and Sreedharan, Sarath and Kambhampati, Subbarao},
  booktitle={Advances in Neural Information Processing Systems},
  volume={36},
  pages={75993--76005},
  year={2023}
}

@article{huang2023hallucination,
  title={A survey on hallucination in large language models: Principles, taxonomy, challenges, and open questions},
  author={Huang, Lei and Yu, Weijiang and Ma, Weitao and Zhong, Weihong and others},
  journal={ACM Transactions on Information Systems},
  volume={43},
  number={2},
  pages={1--55},
  year={2025}
}

@inproceedings{li2023geolm,
  title={{GeoLM}: Empowering language models for geospatially grounded language understanding},
  author={Li, Zekun and Zhou, Wenxuan and Chiang, Yao-Yi and Chen, Muhao},
  booktitle={Proceedings of the 2023 Conference on Empirical Methods in Natural Language Processing},
  pages={5227--5240},
  year={2023}
}

@inproceedings{hoffmann2022training,
  title={Training compute-optimal large language models},
  author={Hoffmann, Jordan and Borgeaud, Sebastian and Mensch, Arthur and Buchatskaya, Elena and others},
  booktitle={Advances in Neural Information Processing Systems},
  pages={30016--30030},
  year={2022}
}

@article{brown2020language,
  title={Language models are few-shot learners},
  author={Brown, Tom and Mann, Benjamin and Ryder, Nick and Subbiah, Melanie and Kaplan, Jared D and Dhariwal, Prafulla and Neelakantan, Arvind and Shyam, Pranav and Sastry, Girish and Askell, Amanda and others},
  journal={Advances in Neural Information Processing Systems},
  volume={33},
  pages={1877--1901},
  year={2020}
}

@article{schick2023toolformer,
  title={Toolformer: Language models can teach themselves to use tools},
  author={Schick, Timo and Dwivedi-Yu, Jane and Dess{\`\i}, Roberto and Raileanu, Roberta and Lomeli, Maria and Hambro, Eric and Zettlemoyer, Luke and Cancedda, Nicola and Scialom, Thomas},
  journal={Advances in Neural Information Processing Systems},
  volume={36},
  pages={68539--68551},
  year={2023}
}

@inproceedings{delling2019fast,
  title={Fast and exact public transit routing with restricted pareto sets},
  author={Delling, Daniel and Dibbelt, Julian and Pajor, Thomas},
  booktitle={Proceedings of the Twenty-First Workshop on Algorithm Engineering and Experiments},
  pages={54--65},
  year={2019}
}

@inproceedings{rajbhandari2020zero,
  title={Zero: Memory optimizations toward training trillion parameter models},
  author={Rajbhandari, Samyam and Rasley, Jeff and Ruwase, Olatunji and He, Yuxiong},
  booktitle={SC20: international conference for high performance computing, networking, storage and analysis},
  pages={1--16},
  year={2020},
  organization={IEEE}
}

@article{de2013unique,
  title={Unique in the crowd: The privacy bounds of human mobility},
  author={De Montjoye, Yves-Alexandre and Hidalgo, C{\'e}sar A and Verleysen, Michel and Blondel, Vincent D},
  journal={Scientific reports},
  volume={3},
  number={1},
  pages={1376},
  year={2013}
}
}

\newpage
\appendix

\section{Data Visualization}
\label{app:visualization}

\begin{figure}[h]
\centering
\includegraphics[width=\textwidth]{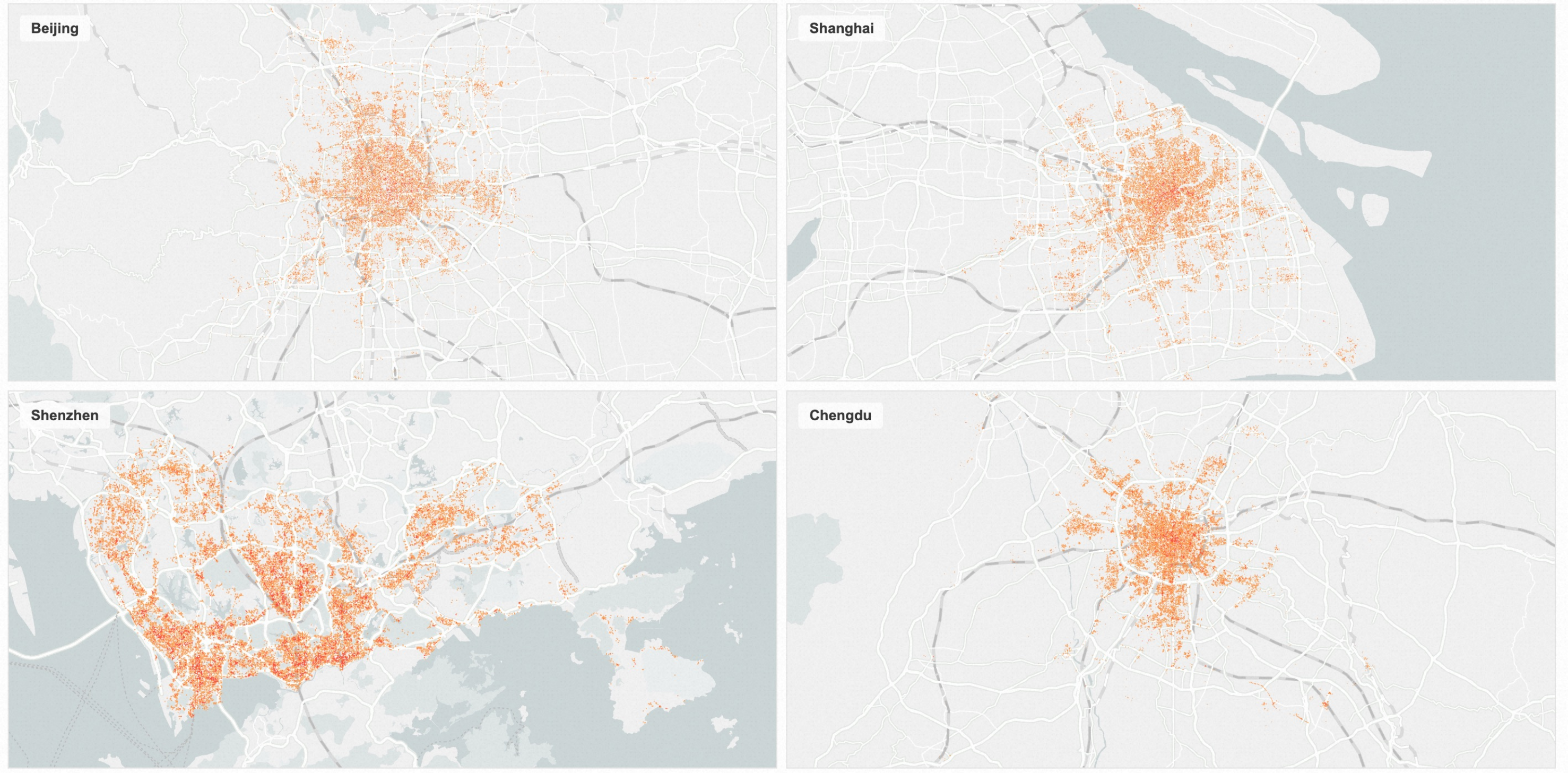}
\caption{Geographic distribution of route planning origins across the four cities. Density reflects real-world transit demand concentration in urban cores.}
\label{fig:heatmap}
\end{figure}

Figure~\ref{fig:heatmap} visualizes the spatial distribution of route planning origins across the four cities. The heatmaps reveal dense coverage in urban cores with natural dispersion toward suburban areas, confirming that the dataset reflects real-world transit demand patterns rather than synthetic or uniformly sampled coordinates.

\section{CPT Corpus Sample}
\label{app:cpt_sample}

The CPT corpus consists of two complementary components: (1)~\textbf{session records} derived from real-world transit planning sessions, each pairing an origin--destination request with candidate routes, and (2)~\textbf{static descriptions} of transit lines and stations. The original corpus is in Chinese. English translations are provided below for readability.

\paragraph{Session Records.}
Each session record follows a \textit{query~$\rightarrow$~route options} structure. The query specifies the city, origin--destination GPS coordinates, and POI names. Each candidate route details the transport mode, line name, per-segment distance and time, fare, boarding/alighting stations with coordinates, and the complete station ID sequence.

\begin{tcolorbox}[colback=gray!3, colframe=black!40, boxrule=0.4pt, arc=2pt, left=6pt, right=6pt, top=4pt, bottom=4pt, fontupper=\small, fontlower=\small, breakable, title={\small\textbf{CPT Session Record}}]
\textbf{Query:} How to get to Yongtai Dongli? \\
\textbf{City:} Beijing \quad \textbf{Origin:} 116.398261, 39.889556 \quad \textbf{Dest:} 116.358844, 40.034880

\tcblower

\textbf{Option 1} \textit{(Subway, 21.3\,km, 55\,min, 5\,yuan)} \\[2pt]
Walk 170\,m (2\,min) to bus-899931 (Zhushikou, 116.398375, 39.889961). \\
Take Subway Line 8 to bus-893448 (Yongtaizhuang, 116.354580, 40.037728), 20.3\,km, 41\,min. \\
Walk 880\,m to destination. \\[2pt]
\textit{Stations:} bus-899931, bus-900093, \ldots, bus-893448.

\vspace{4pt}
\noindent\dashrule
\vspace{4pt}

\textbf{Option 2} \textit{(Bus, 20.8\,km, 1h\,55min, 5\,yuan)} \\[2pt]
Walk 524\,m to bus-899401 (Banzhang Rd, 116.394073, 39.890199). \\
Take Bus 5 to bus-897760 (Madian Bridge S., 116.397834, 39.959271), 10.7\,km, 54\,min. \\
\textit{[Transfer]} Take Bus 625 to bus-892356 (Qinghe, 116.348232, 40.030627), 8.3\,km, 34\,min. \\
Walk 1.4\,km to destination. \\[2pt]
\textit{Stations:} bus-899401, \ldots, bus-897760, [transfer], bus-897760, \ldots, bus-892356.
\end{tcolorbox}

\paragraph{Static Descriptions.}
The corpus also includes structured descriptions of individual transit lines and stations, encoding attributes such as route length, stop count, operating hours, fare policy, coordinates, and connectivity to neighboring stations.

\begin{tcolorbox}[colback=gray!3, colframe=black!40, boxrule=0.4pt, arc=2pt, left=6pt, right=6pt, top=4pt, bottom=4pt, fontupper=\small, fontlower=\small, breakable, title={\small\textbf{CPT Static Descriptions}}]
\textbf{Line Description} \\[2pt]
\textit{Line:} Zhuan-117 (Tiantongyuan North Hub -- Yandan) \\
\textit{Type:} Micro-circulation bus \quad \textit{Length:} 5.3\,km \quad \textit{Stops:} 9 \\
\textit{Hours:} 05:50--22:30 \quad \textit{Fare:} 2\,yuan (flat) \\[2pt]
\textit{Stop sequence:} \\
\#1 Tiantongyuan North Hub (bus-901618) $\rightarrow$
\#2 Tiantong Beiyuan Block\,1 North Gate (bus-902625) $\rightarrow$
\#3 Tiantong Beiyuan Block\,2 North Gate (bus-903364) $\rightarrow$
\#4 Shiziying (bus-904346) $\rightarrow$
\#5 Tiantong Beiyuan Block\,3 North Gate (bus-904981) $\rightarrow$
\#6 Qixing Rd South (bus-905417) $\rightarrow$
\#7 Yanchengyuan Intersection South (bus-905008) $\rightarrow$
\#8 Yanchengyuan (bus-905518) $\rightarrow$
\#9 Yandan (bus-905652).

\tcblower

\textbf{Station Description} \\[2pt]
\textit{Station:} Pinggu No.4 Middle School (bus-1000071) \\
\textit{District:} Pinggu, Beijing \quad \textit{Lines served:} 2 \\
\textit{Coordinates:} (117.089436, 40.130547) \\[2pt]
\textit{Connectivity:} From this station, take Line Ping-2 (Passenger Terminal -- Pinggu No.7 Primary School) to Yuming School (bus-1000202).
\end{tcolorbox}

\section{Benchmark SFT Examples}
\label{app:benchmark_examples}

Each benchmark task uses a standardized prompt--label format. The prompt contains a system instruction describing the task, followed by a user request with origin--destination coordinates. The label is a structured JSON object encoding the expected route. We present representative examples of all three tasks below. The original Chinese text is translated into English for readability.

\paragraph{Optimal Route Generation.}
Given origin--destination coordinates, the model generates a single optimal transit route as a structured JSON object.
\begin{tcolorbox}[colback=gray!3, colframe=black!40, boxrule=0.4pt, arc=2pt, left=6pt, right=6pt, top=4pt, bottom=4pt, fontupper=\small, fontlower=\small, breakable, title={\small\textbf{Prompt \& Label}}]
\textbf{System Prompt} \\[2pt]
You are a public transit route planning expert, familiar with bus and subway networks.

\textit{Task:} \textbf{Given the user's origin and destination coordinates, plan an optimal transit route.}

\textit{Input:} \texttt{query}, \texttt{start} (lng, lat), \texttt{end} (lng, lat), \texttt{city}.

\textit{Output (JSON):} \texttt{line\_sequence}, \texttt{station\_sequence} (with \texttt{[Transfer]}), \texttt{total\_distance}, \texttt{total\_time}, \texttt{total\_fare}, \texttt{start\_transfer\_mode/distance}, \texttt{end\_transfer\_mode/distance}.

\tcblower

\textbf{Prompt} \\[2pt]
\texttt{\{"query": "How to get to BIT (South Gate)?", "start": "116.395310,39.963284",} \\
\texttt{~"end": "116.316356,39.957053", "city": "Beijing"\}}

\vspace{6pt}
\textbf{Label} \\[2pt]
\texttt{\{"line\_sequence": ["Subway Line 12", "Subway Line 4"],} \\
\texttt{~"station\_sequence": ["bus-899733", "bus-898054", ..., "bus-888019",} \\
\texttt{~~"[Transfer]", "bus-887631", "bus-887990"],} \\
\texttt{~"total\_distance": "9.2 km", "total\_time": "49 min", "total\_fare": "4 yuan",} \\
\texttt{~"start\_transfer\_mode": "walking", "start\_transfer\_distance": "896 m",} \\
\texttt{~"end\_transfer\_mode": "walking", "end\_transfer\_distance": "759 m"\}}
\end{tcolorbox}

\paragraph{Preference-Aware Planning.}
In addition to coordinates, the prompt includes a user preference constraint. The model must produce a route that satisfies the stated preference while maintaining overall quality.
\begin{tcolorbox}[colback=gray!3, colframe=black!40, boxrule=0.4pt, arc=2pt, left=6pt, right=6pt, top=4pt, bottom=4pt, fontupper=\small, fontlower=\small, breakable, title={\small\textbf{Prompt \& Label}}]
\textbf{System Prompt} \\[2pt]
You are a public transit route planning expert, familiar with bus and subway networks.

\textit{Task:} \textbf{Given the user's origin and destination coordinates along with a stated preference, plan a transit route that satisfies the preference while maintaining overall route quality.}

\textit{Input:} \texttt{query} (with preference), \texttt{start} (lng, lat), \texttt{end} (lng, lat), \texttt{city}.

\textit{Output (JSON):} \texttt{line\_sequence}, \texttt{station\_sequence} (with \texttt{[Transfer]}), \texttt{total\_distance}, \texttt{total\_time}, \texttt{total\_fare}, \texttt{start\_transfer\_mode/distance}, \texttt{end\_transfer\_mode/distance}.

\tcblower

\textbf{Prompt} \\[2pt]
\texttt{\{"query": "How to get to Jingshan Park by bus? Bus-first.",} \\
\texttt{~"start": "116.421180,39.880405", "end": "116.396551,39.925875", "city": "Beijing"\}}

\vspace{6pt}
\textbf{Label} \\[2pt]
\texttt{\{"line\_sequence": ["Bus 128 (Xiaocunqiao N.--Beijing No.39 Middle School)"],} \\
\texttt{~"station\_sequence": ["bus-903008", "bus-902952", "bus-902743", "bus-902705",} \\
\texttt{~~"bus-902688", "bus-902629", "bus-902600", "bus-902571", "bus-902555",} \\
\texttt{~~"bus-902495", "bus-902472", "bus-901722", "bus-900713", "bus-899614"],} \\
\texttt{~"total\_distance": "7.0 km", "total\_time": "53 min", "total\_fare": "2 yuan",} \\
\texttt{~"start\_transfer\_mode": "walking", "start\_transfer\_distance": "143 m",} \\
\texttt{~"end\_transfer\_mode": "walking", "end\_transfer\_distance": "145 m"\}}
\end{tcolorbox}

\paragraph{Multi-Route Generation.}
The model produces three diverse transit routes for a single origin--destination pair. Each route should adopt a different mode or line combination to offer meaningful travel alternatives, while maintaining overall route quality.
\begin{tcolorbox}[colback=gray!3, colframe=black!40, boxrule=0.4pt, arc=2pt, left=6pt, right=6pt, top=4pt, bottom=4pt, fontupper=\small, fontlower=\small, breakable, title={\small\textbf{Prompt \& Label}}]
\textbf{System Prompt} \\[2pt]
You are a public transit route planning expert, familiar with bus and subway networks.

\textit{Task:} \textbf{Given origin--destination coordinates and city, plan three diverse transit routes using different mode or line combinations.}

\textit{Input:} \texttt{query}, \texttt{start} (lng, lat), \texttt{end} (lng, lat), \texttt{city}.

\textit{Output (JSON):} Three route objects (\texttt{first}, \texttt{second}, \texttt{third}), each containing \texttt{route\_tag}, \texttt{line\_sequence}, \texttt{station\_sequence}, \texttt{total\_distance}, \texttt{total\_time}, \texttt{total\_fare}, \texttt{start/end\_transfer\_mode/distance}.

\tcblower

\textbf{Prompt} \\[2pt]
\texttt{\{"query": "How to get from Tongji Univ.\ to Shanghai Pudong Intl.\ Airport} \\
\texttt{~by public transit? Give me a few options.",} \\
\texttt{~"start": "121.513244,31.282003", "end": "121.808681,31.142267",} \\
\texttt{~"city": "Shanghai"\}}

\vspace{6pt}
\textbf{Label} \\[4pt]
\texttt{\{"first": \{"route\_tag": "Subway",} \\
\texttt{~~"line\_sequence": ["Subway Line 18", "Subway Line 2"],} \\
\texttt{~~"station\_sequence": ["bus-1670095", \ldots, "bus-1677113",} \\
\texttt{~~~"[Transfer]", "bus-1677227", \ldots, "bus-1701975"],} \\
\texttt{~~"total\_distance": "41.1 km", "total\_time": "1h 22min", "total\_fare": "7 yuan",} \\
\texttt{~~"start\_transfer\_mode": "walking", "start\_transfer\_distance": "169 m",} \\
\texttt{~~"end\_transfer\_mode": "walking", "end\_transfer\_distance": "22 m"\},} \\[3pt]
\texttt{~"second": \{"route\_tag": "Bus+Subway",} \\
\texttt{~~"line\_sequence": ["Subway Line 18", "Maglev Line"],} \\
\texttt{~~"station\_sequence": [\ldots], \ldots\},} \\[3pt]
\texttt{~"third": \{"route\_tag": "Bus",} \\
\texttt{~~"line\_sequence": ["Airport Express Line 4"],} \\
\texttt{~~"station\_sequence": [\ldots], \ldots\}\}}
\end{tcolorbox}

\section{Evaluation Metrics}
\label{app:metrics}

This appendix provides the formal definitions of all evaluation metrics introduced in Section~\ref{benchmark}. The metrics cover four complementary dimensions: \textit{Connectivity} verifies structural correctness of the predicted station sequence; \textit{Access Feasibility} checks whether first/last-mile access is physically plausible; \textit{Route Overlap} measures structural match between predicted and ground-truth routes; and \textit{Numeric Field Accuracy} assesses the accuracy of predicted numeric fields. In addition, Preference-Aware Planning and Multi-Route Generation define task-specific metrics for preference compliance and route diversity, respectively. Table~\ref{tab:metric_abbr} provides a quick-reference summary of all metric abbreviations.

\begin{table}[h]
\centering
\small
\caption{Evaluation metrics and abbreviations.}
\label{tab:metric_abbr}
\renewcommand{\arraystretch}{1.05}
\resizebox{\textwidth}{!}{
\begin{tabular}{@{}ll p{0.52\textwidth}@{}}
\toprule
\textbf{Metric} & \textbf{Abbr.} & \textbf{Description} \\
\midrule
\multicolumn{3}{l}{\textit{Evaluation Dimensions}} \\[2pt]
Connectivity & Conn & Consecutive stations reachable in the transit network \\
\cdashline{1-3}[0.4pt/3pt]
Access Feasibility & & First/last-mile access plausibility \\
\quad Station Grounding & SG & First/last station within mode-specific distance of OD \\
\quad Distance Plausibility & DP & Predicted access distance within geometric plausibility bounds \\
\cdashline{1-3}[0.4pt/3pt]
Route Overlap & & Structural match between predicted and label routes \\
\quad Line Overlap & LO & IoU of full line sets including access segments \\
\quad Station Sequence Overlap & SSO & IoU of station ID sets \\
\quad Route Exact Match & REM & Fraction of samples with LO\,=\,1 and SSO\,=\,1 \\
\cdashline{1-3}[0.4pt/3pt]
Numeric Field Accuracy & & Accuracy of predicted numeric fields \\
\quad Estimation Accuracy & EA & Avg.\ pass rate over distance, time, and fare \\
\quad Mean Absolute Percentage Error & MAPE & Avg.\ relative error over distance, time, and fare \\
\midrule
\multicolumn{3}{l}{\textit{Task-specific Metrics}} \\[2pt]
Preference Compliance & PC & Stated preference satisfied \\
Route Diversity & RD & Pairwise line-set dissimilarity \\
\bottomrule
\end{tabular}
}
\end{table}

\subsection{Connectivity}

Connectivity is the first evaluation dimension, verifying that the predicted station sequence forms a structurally valid path in the transit network. A predicted route is \emph{connected} if and only if every consecutive station pair $(s_i, s_{i+1})$ in the generated sequence is reachable, either on the same line or via a valid transfer recorded in the city's transfer table. Connectivity is reported as the percentage of test samples whose predicted routes are fully connected:
\begin{equation}
\text{Conn} = \frac{1}{N}\sum_{i=1}^{N} \mathbf{1}\!\left[\,\forall\, 1 \leq j < L^{(i)},\; (s_j^{(i)}, s_{j+1}^{(i)}) \in \mathcal{E}\,\right]
\end{equation}
where $L^{(i)}$ is the length of the predicted station sequence for sample $i$ and $\mathcal{E}$ is the set of station pairs that are adjacent on a shared line or connected via an inter-line transfer. Connectivity serves as a prerequisite: all subsequent metrics except task-specific ones (PC, RD) are computed only on connected samples. Let $\mathcal{C}$ denote the set of connected test samples.

\subsection{Access Feasibility}

Access Feasibility is the second evaluation dimension. While Connectivity verifies reachability among intermediate stations, this metric validates the first/last-mile segments, jointly ensuring that the entire origin-to-destination path is both connected and accessible. For each access segment, let $d_{\text{geo}}$ denote the straight-line (Haversine) distance between the origin/destination and the predicted boarding/alighting station, and let $d_{\text{pred}}$ denote the predicted access distance. Let $a_{\mathrm{s}}^{(i)}$ and $a_{\mathrm{e}}^{(i)}$ denote the start and end access segments of sample $i$.

\paragraph{Station Grounding.} This metric evaluates whether the model can implicitly map raw GPS coordinates to nearby transit stations without any explicit coordinate-to-station lookup module. A high pass rate indicates that the model has learned spatial grounding purely from training data. The straight-line distance between the origin/destination and the predicted boarding/alighting station must not exceed a mode-specific threshold:
\begin{center}
\small
\begin{tabular}{l@{\hspace{2em}}c}
\toprule
Access Mode & Threshold $\tau_m$ \\
\midrule
Walking & 3\,km \\
Cycling & 5\,km \\
Taxi & 10\,km \\
\bottomrule
\end{tabular}
\end{center}
\noindent The pass rate is:
\begin{equation}
\text{SG} = \frac{1}{|\mathcal{C}|}\sum_{i\in\mathcal{C}} \mathbf{1}\!\left[\,d_{\text{geo}}(a_{\mathrm{s}}^{(i)}) \leq \tau_{m_{\mathrm{s}}} \;\wedge\; d_{\text{geo}}(a_{\mathrm{e}}^{(i)}) \leq \tau_{m_{\mathrm{e}}}\,\right]
\end{equation}
where $\tau_{m}$ is the mode-specific threshold from the table above.

\paragraph{Distance Plausibility.} While Station Grounding validates spatial proximity, this metric further verifies that the predicted access distance is physically realistic rather than a hallucinated value. A plausible access distance must satisfy two conditions:
\begin{itemize}[leftmargin=2em, itemsep=2pt]
\item $d_{\text{pred}} \geq d_{\text{geo}}$ \quad (predicted distance must not fall below the geometric lower bound)
\item $d_{\text{pred}} \leq 3 \cdot d_{\text{geo}}$ \quad (predicted distance not unreasonably large)
\end{itemize}
\noindent Let $\phi(a) = \mathbf{1}[d_{\text{geo}}(a) \leq d_{\text{pred}}(a) \leq 3 \cdot d_{\text{geo}}(a)]$. The pass rate is:
\begin{equation}
\text{DP} = \frac{1}{|\mathcal{C}|}\sum_{i\in\mathcal{C}} \phi(a_{\mathrm{s}}^{(i)}) \cdot \phi(a_{\mathrm{e}}^{(i)})
\end{equation}
The upper-bound factor of 3 is empirically set as a plausibility threshold: real-world access distances rarely exceed three times the straight-line distance, so predictions beyond this bound are considered implausible.

\subsection{Route Overlap}
Route Overlap is the third evaluation dimension, quantifying the structural match between a predicted route and its ground-truth counterpart. For Optimal Route Generation and Preference-Aware Planning, the single predicted route is directly compared against the label. For Multi-Route Generation, the first predicted route is used for comparison, as the task requires the first output route to be the ground-truth route, and the training and evaluation data are constructed accordingly. Both Line Overlap and Station Sequence Overlap are defined via Intersection-over-Union (IoU):
\begin{equation}
\text{IoU}(A, B) = \frac{|A \cap B|}{|A \cup B|}
\end{equation}
For \textit{Line Overlap} ($\text{LO}$), $A$ and $B$ are the full line sets of the predicted and ground-truth routes, including first/last-mile access segments (e.g., cycling, taxi). For \textit{Station Sequence Overlap} ($\text{SSO}$), $A$ and $B$ are the corresponding station ID sets.

\textit{Route Exact Match} ($\text{REM}$) measures the fraction of connected samples whose predicted line and station ID sets exactly match those of the reference route:
\begin{equation}
\text{REM} = \frac{1}{|\mathcal{C}|}\sum_{i\in\mathcal{C}} \mathbf{1}\!\left[\,\text{LO}^{(i)} = 1 \;\wedge\; \text{SSO}^{(i)} = 1\,\right]
\end{equation}

\subsection{Numeric Field Accuracy}

Numeric Field Accuracy is the fourth evaluation dimension, measuring how accurately the model predicts route-level numeric attributes. Let $\mathcal{F} = \{\text{distance}, \text{time}, \text{fare}\}$ denote the set of route-level numeric fields. Evaluation is restricted to samples that achieve Route Exact Match, i.e., both $\text{LO} = 1$ and $\text{SSO} = 1$, as only under this condition do the ground-truth numeric fields constitute valid references. Let $\mathcal{M}$ denote this set of matched samples.

\textbf{\textit{Estimation Accuracy (EA).}}
For each field $f \in \mathcal{F}$, a prediction passes if it satisfies \textit{either} a relative tolerance \textit{or} an absolute tolerance:
\begin{equation}
\text{EA}_f = \frac{1}{|\mathcal{M}|}\sum_{i \in \mathcal{M}} \mathbf{1}\!\left[\frac{|\hat{y}_f^{(i)} - y_f^{(i)}|}{y_f^{(i)}} \leq 10\% \;\;\text{or}\;\; |\hat{y}_f^{(i)} - y_f^{(i)}| \leq \epsilon_f\right]
\end{equation}
where $\epsilon_{\text{time}} = 5\,\text{min}$, $\epsilon_{\text{distance}} = 500\,\text{m}$, and $\epsilon_{\text{fare}} = 1\,\text{CNY}$. The reported EA is the average across all fields: $\text{EA} = \frac{1}{|\mathcal{F}|}\sum_{f \in \mathcal{F}} \text{EA}_f$. The dual-tolerance design ensures that predictions with small absolute error but inflated relative error (due to small denominators) are not penalized, and likewise for predictions with small relative error but large absolute difference.

\textbf{\textit{Mean Absolute Percentage Error (MAPE).}} While EA provides a binary pass/fail judgment, MAPE quantifies the continuous error magnitude. The per-field MAPE is:
\begin{equation}
\text{MAPE}_f = \frac{1}{|\mathcal{M}|}\sum_{i \in \mathcal{M}} \frac{|\hat{y}_f^{(i)} - y_f^{(i)}|}{y_f^{(i)}}
\end{equation}
The reported MAPE is the average across all fields: $\text{MAPE} = \frac{1}{|\mathcal{F}|}\sum_{f \in \mathcal{F}} \text{MAPE}_f$.

\subsection{Preference Compliance}

Preference Compliance is a task-specific metric for Preference-Aware Planning that evaluates whether the model can generate routes adhering to an explicit user preference. Let $r_{\text{gt}}$ denote the ground-truth route. For each preference type, compliance is determined by a hard rule:

\begin{center}
\small
\begin{tabular}{lp{8cm}}
\toprule
\textbf{Preference} & \textbf{Compliance Rule} \\
\midrule
Subway-first & The predicted route contains at least one subway segment. \\
Bus-first & The predicted route contains at least one bus segment. \\
Fewer transfers & $N_{\text{transfers}}^{\text{pred}} \leq N_{\text{transfers}}^{\text{gt}}$ \\
Shortest time & $T^{\text{pred}} \leq \alpha \cdot T^{\text{gt}}$, where $\alpha = 1.1$ \\
\bottomrule
\end{tabular}
\end{center}

\noindent The tolerance factor $\alpha$ for shortest time accounts for the inherent imprecision of predicted numeric values. Preference Compliance is reported as the percentage of test samples whose predicted routes satisfy the corresponding rule. Routes that strictly satisfy a preference may have poor overall quality, so the ground-truth labels prioritize route quality and do not always comply with the hard rule.

\subsection{Route Diversity}

Route Diversity is a task-specific metric for Multi-Route Generation that evaluates whether the model can produce structurally distinct alternatives rather than near-duplicate routes. For each sample, the pairwise dissimilarity over all $\binom{3}{2} = 3$ route pairs is:
\begin{equation}
\text{RD}^{(k)} = \frac{1}{3}\sum_{(i,j)}(1 - \text{IoU}(\mathcal{L}_i^{(k)}, \mathcal{L}_j^{(k)}))
\end{equation}
where $\mathcal{L}_i^{(k)}$ is the full line set of route $i$ in sample $k$, including cycling and taxi access segments. The reported metric is the average over all $N$ test samples: $\text{RD} = \frac{1}{N}\sum_{k=1}^{N}\text{RD}^{(k)}$. Values range from 0 (all routes identical) to 1 (no shared lines). Route Diversity should be evaluated jointly with the four evaluation dimensions, as maximizing diversity alone may sacrifice route quality. The two metrics together capture whether the model produces varied yet practically viable alternatives.

\section{Hyperparameters}
\label{app:hyperparams}

Table~\ref{tab:hyperparams} summarizes all hyperparameters used in training and inference. During CPT, all sequences are packed to the fixed sequence length with no padding, so the training budget is measured in steps rather than epochs. During SFT, sequences are not packed and instead processed at their natural length with standard padding; the loss is computed only on the response tokens, with prompt tokens masked. Both stages use DeepSpeed ZeRO-3 \cite{rajbhandari2020zero} for distributed training. Greedy decoding with a fixed seed is used for all benchmark evaluations to ensure deterministic and reproducible outputs.

\begin{table}[t]
\centering
\caption{Hyperparameters for training and inference.}
\label{tab:hyperparams}
\small
\setlength{\tabcolsep}{8pt}
\begin{tabular}{@{}lcc@{}}
\toprule
\textbf{Hyperparameter} & \textbf{CPT} & \textbf{SFT} \\
\midrule
\multicolumn{3}{l}{\textit{Model}} \\[2pt]
\quad Backbone & \multicolumn{2}{c}{Qwen3-Base (0.6B / 1.7B / 4B)} \\
\quad Initialization & Base weights & CPT checkpoint \\
\quad Added tokens & \multicolumn{2}{c}{120,845 station IDs} \\
\midrule
\multicolumn{3}{l}{\textit{Optimization}} \\[2pt]
\quad Learning rate & 2e-5 & 2e-5 \\
\quad LR scheduler & Cosine & Cosine \\
\quad Optimizer & AdamW & AdamW \\
\quad Warmup steps & 500 & 15 \\
\quad Weight decay & 0.01 & 0.01 \\
\quad Precision & bf16 & bf16 \\
\quad Distributed strategy & \multicolumn{2}{c}{DeepSpeed ZeRO-3} \\
\quad Loss masking & Full sequence & Response only \\
\midrule
\multicolumn{3}{l}{\textit{Data \& Compute}} \\[2pt]
\quad Sequence length & 4,096 (packed) & Dynamic \\
\quad Per-device batch size & 4 & 8 \\
\quad Gradient accumulation & 4 & 4 \\
\quad Number of devices & 64 PPUs (96\,GB) & 8 PPUs (96\,GB) \\
\quad Effective batch size & 1,024 & 256 \\
\quad Training budget & 15,000 steps & 1 epoch \\
\midrule
\multicolumn{3}{l}{\textit{Inference}} \\[2pt]
\quad Decoding strategy & \multicolumn{2}{c}{Greedy} \\
\quad Max new tokens & \multicolumn{2}{c}{4,096} \\
\quad Input cutoff length & \multicolumn{2}{c}{2,048} \\
\quad Seed & \multicolumn{2}{c}{1} \\
\bottomrule
\end{tabular}
\end{table}

\section{Additional Experiments}
\label{app:additional_exp}

\subsection{CPT Training Dynamics}
\label{app:cpt_loss}

\begin{figure}[t]
\centering
\includegraphics[width=0.85\textwidth]{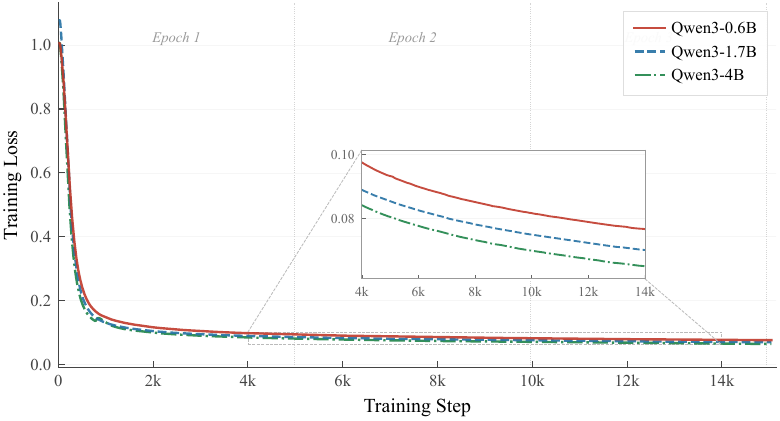}
\caption{CPT training loss curves for Qwen3-0.6B, Qwen3-1.7B, and Qwen3-4B over $\approx$\,15k steps. The inset magnifies steps 4k--14k to highlight the sustained loss reduction and the capacity gap across model sizes.}
\label{fig:cpt_loss}
\end{figure}

Figure~\ref{fig:cpt_loss} plots the CPT training loss for the three backbone models over approximately 15k steps ($\approx$\,3 epochs). All models drop from $>$\,1.0 to approximately 0.1 within the first 2k steps, indicating that domain-specific token distributions are learned early regardless of model capacity. The inset (steps 4k--14k) reveals a stable ordering Qwen3-4B $<$ Qwen3-1.7B $<$ Qwen3-0.6B in loss, consistent with downstream performance in Tables~\ref{tab:task1_results}--\ref{tab:task3_results}. Loss continues to decrease through Epochs~2 and~3 (e.g., 4B: 0.084\,$\to$\,0.070), suggesting that later epochs further consolidate transit-domain knowledge. Wall-clock training time on 64 PPUs is approximately 6 days for Qwen3-4B, 3 days for Qwen3-1.7B, and 1.5 days for Qwen3-0.6B, scaling approximately with model size.

\subsection{Single-City vs.\ Multi-City CPT}
\label{app:single_city}

Our framework introduces every station ID as a new token in the vocabulary. Scaling from one city to four increases the station vocabulary from 38{,}792 to 120{,}845, a 3.1$\times$ expansion. Because the total CPT data volume is held constant, each Beijing station receives roughly one-third as many training examples under the multi-city setting. This ablation examines whether the resulting token-level sparsity degrades per-city performance, and whether cross-city training provides compensating knowledge transfer. To this end, we train a single-city model on Beijing data only, using the same Qwen3-4B base and identical CPT data volume as the four-city model. Both models are evaluated on the Beijing test set of 10{,}000 samples for Optimal Route Generation. Table~\ref{tab:single_city} reports the comparison.

\begin{table}[t]
\centering
\caption{Single-city vs.\ multi-city CPT on \textbf{Optimal Route Generation}, evaluated on the Beijing test set with 10{,}000 samples. The single-city model is trained on Beijing data only, while the multi-city model covers all four cities. Both use the same total CPT data volume and identical SFT data. $\Delta$ denotes the change from Beijing-only to four-city. $\uparrow$\,/\,$\downarrow$ indicate higher/lower is better. \textbf{Bold}: better of the two.}
\label{tab:single_city}
\small
\setlength{\tabcolsep}{6pt}
\renewcommand{\arraystretch}{1.05}
\scalebox{1.05}{
\begin{tabular}{@{}l@{\enspace}c ccc@{}}
\toprule
\textbf{Metric} & & \textbf{Beijing-Only} & \textbf{Four-City} & \textbf{$\Delta$} \\
\midrule
Connectivity & $\uparrow$ & \textbf{95.8\%} & 95.5\% & $-$0.3 \\
\cdashline{1-5}[0.4pt/3pt]
Station Grounding & $\uparrow$ & 97.5\% & \textbf{97.8\%} & \textcolor{ForestGreen}{+0.3} \\
Distance Plausibility & $\uparrow$ & \textbf{89.7\%} & 88.8\% & $-$0.9 \\
\cdashline{1-5}[0.4pt/3pt]
Line Overlap & $\uparrow$ & \textbf{0.853} & 0.845 & $-$0.008 \\
Station Sequence Overlap & $\uparrow$ & \textbf{0.853} & 0.839 & $-$0.014 \\
Route Exact Match & $\uparrow$ & \textbf{74.0\%} & 70.5\% & $-$3.5 \\
\cdashline{1-5}[0.4pt/3pt]
Estimation Accuracy & $\uparrow$ & 97.1\% & \textbf{97.6\%} & \textcolor{ForestGreen}{+0.5} \\
MAPE & $\downarrow$ & \textbf{1.30\%} & 1.63\% & +0.33 \\
\bottomrule
\end{tabular}
}
\end{table}

Despite a 3.1$\times$ vocabulary expansion and proportionally fewer per-station training examples, the four-city model trails the Beijing-only model by only 3.5 percentage points in Route Exact Match. This confirms that token-level sparsity introduced by city scaling does not cause significant performance degradation. Station Grounding and Estimation Accuracy are slightly higher under multi-city training, suggesting that shared spatial patterns across cities provide positive knowledge transfer that partially compensates for reduced per-station coverage. The minor increase in MAPE and decrease in route-level overlap metrics represent a modest cost of distributing the same data budget across a larger station vocabulary. These results validate that the proposed framework scales gracefully to additional cities, directly supporting the future direction of extending coverage to broader geographies.

\subsection{Effect of Continual Pre-Training}
\label{app:cpt_ablation}

To isolate the contribution of continual pre-training, we construct an SFT-only baseline that bypasses the CPT stage and instead trains the base Qwen3-4B model directly on the same volume of session data used by CPT-25\%, reformatted as supervised fine-tuning examples. This ensures that the comparison controls for total data volume rather than merely removing a training stage. We additionally include CPT-25\%, CPT-100\%, and the multi-task 4B-Joint model. All variants share identical station token vocabulary and are evaluated on the same test set. Each configuration is evaluated under both standard text input and GPS-only input to probe the robustness of the learned representations. Table~\ref{tab:cpt_ablation} reports the results.

\begin{table}[t]
\centering
\caption{Effect of continual pre-training on \textbf{Optimal Route Generation} with 10{,}000 test samples. SFT-only bypasses the CPT stage and trains on the same session data volume as CPT-25\%, reformatted as SFT examples. All variants share identical station token vocabulary and test set. Each configuration is evaluated under standard text and GPS-only input. $\uparrow$\,/\,$\downarrow$ indicate higher/lower is better. \textbf{Bold}: best; \underline{underline}: second best.}
\label{tab:cpt_ablation}
\small
\setlength{\tabcolsep}{3pt}
\renewcommand{\arraystretch}{1.05}
\scalebox{1.0}{
\begin{tabular}{@{}l@{\enspace}c cccc@{}}
\toprule
\textbf{Metric} & & \textbf{SFT-only} & \textbf{CPT-25\%} & \textbf{CPT-100\%} & \textbf{4B-Joint} \\
\midrule
\multicolumn{6}{@{}l}{\textit{Standard input}} \\
Connectivity & $\uparrow$ & \underline{97.3\%} & 95.9\% & 97.0\% & \textbf{97.9\%} \\
Station Grounding & $\uparrow$ & 97.2\% & 97.7\% & \underline{98.5\%} & \textbf{98.9\%} \\
Distance Plausibility & $\uparrow$ & 90.7\% & 87.4\% & \underline{91.0\%} & \textbf{92.9\%} \\
Line Overlap & $\uparrow$ & 0.820 & 0.811 & \underline{0.828} & \textbf{0.835} \\
Station Sequence Overlap & $\uparrow$ & 0.835 & 0.816 & \underline{0.838} & \textbf{0.847} \\
Route Exact Match & $\uparrow$ & \textbf{74.9\%} & 65.6\% & 71.0\% & \underline{73.7\%} \\
Estimation Accuracy & $\uparrow$ & 98.1\% & 97.8\% & \underline{98.5\%} & \textbf{98.6\%} \\
MAPE & $\downarrow$ & 1.35\% & 1.88\% & \underline{1.33\%} & \textbf{1.30\%} \\
\midrule
\multicolumn{6}{@{}l}{\textit{GPS-only input}} \\
Connectivity & $\uparrow$ & 97.0\% & 95.8\% & \underline{97.2\%} & \textbf{98.0\%} \\
Station Grounding & $\uparrow$ & 96.8\% & 97.7\% & \underline{98.3\%} & \textbf{98.8\%} \\
Distance Plausibility & $\uparrow$ & 88.6\% & 87.0\% & \underline{90.4\%} & \textbf{92.6\%} \\
Line Overlap & $\uparrow$ & 0.764 & 0.807 & \underline{0.821} & \textbf{0.827} \\
Station Sequence Overlap & $\uparrow$ & 0.779 & 0.810 & \underline{0.832} & \textbf{0.839} \\
Route Exact Match & $\uparrow$ & 66.1\% & 65.1\% & \underline{70.4\%} & \textbf{72.9\%} \\
Estimation Accuracy & $\uparrow$ & 76.3\% & 96.8\% & \underline{97.6\%} & \textbf{98.1\%} \\
MAPE & $\downarrow$ & 4.96\% & 2.24\% & \underline{1.75\%} & \textbf{1.52\%} \\
\midrule
\multicolumn{6}{@{}l}{\textit{Degradation from text to GPS-only}} \\
Route Exact Match & & \textcolor{red}{$-$8.8} & $-$0.5 & $-$0.6 & $-$0.8 \\
Estimation Accuracy & & \textcolor{red}{$-$21.8} & $-$1.0 & $-$0.9 & $-$0.5 \\
MAPE & & \textcolor{red}{+3.61} & +0.36 & +0.42 & +0.22 \\
\bottomrule
\end{tabular}
}
\end{table}

Under standard text input, SFT-only achieves the highest Route Exact Match at 74.9\%, surpassing even CPT-100\% at 71.0\% and 4B-Joint at 73.7\%. However, the GPS-only evaluation reverses this ranking. When textual cues are removed, SFT-only Route Exact Match drops by 8.8 percentage points, Estimation Accuracy collapses by 21.8 percentage points, and MAPE nearly quadruples from 1.35\% to 4.96\%. All CPT-based models remain nearly unchanged, with Route Exact Match declining by at most 0.8 percentage points. The 4B-Joint model exhibits the smallest EA degradation at only 0.5 percentage points and the lowest GPS-only MAPE at 1.52\%.

This asymmetry reveals that the two training strategies produce representations of fundamentally different nature. The SFT-only model relies disproportionately on textual cues in the query to infer spatial context, yielding strong performance when such cues are available but degrading sharply without them. CPT forces the model to acquire spatial representations from raw transit network data before any task-specific supervision. The resulting representations encode network topology and spatial relationships independently of prompt templates, making them inherently task-agnostic. The 4B-Joint results confirm this directly. Built on CPT-derived representations, Joint training achieves the best GPS-only performance across all metrics, demonstrating that CPT-stage spatial knowledge transfers to multi-task settings with no negative interference.

\begin{table}[t]
\centering
\caption{Data scaling on \textbf{Preference-Aware Planning}: Qwen3-4B trained with varying CPT session data fractions (6.25\%, 12.5\%, 25\%, 50\%) with 100\% as the reference. All variants share identical static descriptions and SFT data. $\uparrow$\,/\,$\downarrow$ indicate higher/lower is better. \textbf{Bold}: best; \underline{underline}: second best.}
\label{tab:scaling_task2}
\small
\setlength{\tabcolsep}{3pt}
\renewcommand{\arraystretch}{1.05}
\scalebox{1.05}{
\begin{tabular}{@{}l@{\enspace}c ccccc@{}}
\toprule
\textbf{Metric} & & \textbf{4B-6.25\%} & \textbf{4B-12.5\%} & \textbf{4B-25\%} & \textbf{4B-50\%} & \textbf{4B-100\%} \\
\midrule
Connectivity & $\uparrow$ & 83.3\% & 86.9\% & 90.1\% & \underline{92.6\%} & \textbf{93.2\%} \\
\cdashline{1-7}[0.4pt/3pt]
Station Grounding & $\uparrow$ & 86.0\% & 92.2\% & 94.2\% & \underline{95.7\%} & \textbf{96.5\%} \\
Distance Plausibility & $\uparrow$ & 59.6\% & 73.2\% & 77.4\% & \underline{82.5\%} & \textbf{84.6\%} \\
\cdashline{1-7}[0.4pt/3pt]
Line Overlap & $\uparrow$ & 0.608 & 0.669 & 0.683 & \underline{0.696} & \textbf{0.705} \\
Station Sequence Overlap & $\uparrow$ & 0.598 & 0.667 & 0.683 & \underline{0.704} & \textbf{0.716} \\
Route Exact Match & $\uparrow$ & 32.6\% & 41.5\% & 43.9\% & \underline{47.8\%} & \textbf{50.4\%} \\
\cdashline{1-7}[0.4pt/3pt]
Estimation Accuracy & $\uparrow$ & 85.9\% & 90.1\% & 90.9\% & \underline{92.2\%} & \textbf{92.5\%} \\
MAPE & $\downarrow$ & 3.67\% & 2.85\% & 2.49\% & \underline{2.12\%} & \textbf{2.05\%} \\
\cdashline{1-7}[0.4pt/3pt]
Preference Compliance & $\uparrow$ & 87.3\% & 87.9\% & \underline{88.8\%} & 88.7\% & \textbf{89.8\%} \\
\bottomrule
\end{tabular}
}
\end{table}

\begin{table}[t]
\centering
\caption{Data scaling on \textbf{Multi-Route Generation}: Qwen3-4B trained with varying CPT session data fractions (6.25\%, 12.5\%, 25\%, 50\%) with 100\% as the reference. All variants share identical static descriptions and SFT data. $\uparrow$\,/\,$\downarrow$ indicate higher/lower is better. \textbf{Bold}: best; \underline{underline}: second best.}
\label{tab:scaling_task3}
\small
\setlength{\tabcolsep}{3pt}
\renewcommand{\arraystretch}{1.05}
\scalebox{1.05}{
\begin{tabular}{@{}l@{\enspace}c ccccc@{}}
\toprule
\textbf{Metric} & & \textbf{4B-6.25\%} & \textbf{4B-12.5\%} & \textbf{4B-25\%} & \textbf{4B-50\%} & \textbf{4B-100\%} \\
\midrule
Connectivity & $\uparrow$ & 91.6\% & 93.9\% & 94.9\% & \underline{95.9\%} & \textbf{96.3\%} \\
\cdashline{1-7}[0.4pt/3pt]
Station Grounding & $\uparrow$ & 92.4\% & 96.2\% & 97.0\% & \underline{97.7\%} & \textbf{98.0\%} \\
Distance Plausibility & $\uparrow$ & 72.0\% & 82.6\% & 84.8\% & \underline{87.9\%} & \textbf{90.1\%} \\
\cdashline{1-7}[0.4pt/3pt]
Line Overlap & $\uparrow$ & 0.671 & 0.740 & 0.756 & \underline{0.776} & \textbf{0.782} \\
Station Sequence Overlap & $\uparrow$ & 0.669 & 0.743 & 0.761 & \underline{0.785} & \textbf{0.794} \\
Route Exact Match & $\uparrow$ & 42.5\% & 55.1\% & 57.9\% & \underline{61.0\%} & \textbf{64.5\%} \\
\cdashline{1-7}[0.4pt/3pt]
Estimation Accuracy & $\uparrow$ & 90.5\% & 96.3\% & 97.4\% & \underline{97.8\%} & \textbf{98.0\%} \\
MAPE & $\downarrow$ & 3.64\% & 2.19\% & 1.90\% & \underline{1.62\%} & \textbf{1.45\%} \\
\cdashline{1-7}[0.4pt/3pt]
Route Diversity & $\uparrow$ & 0.507 & 0.531 & 0.534 & \underline{0.540} & \textbf{0.545} \\
\bottomrule
\end{tabular}
}
\end{table}

\begin{table}[t]
\centering
\caption{GPS-only ablation on our domain-specific models for \textbf{Preference-Aware Planning} with 10{,}000 test samples across four cities. All textual cues are removed and only origin--destination GPS coordinates with preference type are provided as input. $\uparrow$\,/\,$\downarrow$ indicate higher/lower is better. \textbf{Bold}: best; \underline{underline}: second best.}
\label{tab:gps_task2}
\small
\setlength{\tabcolsep}{3pt}
\renewcommand{\arraystretch}{1.05}
\scalebox{1.05}{
\begin{tabular}{@{}l@{\enspace}c ccccc@{}}
\toprule
\textbf{Metric} & & \textbf{Qwen3-0.6B} & \textbf{Qwen3-1.7B} & \textbf{Qwen3-4B-25} & \textbf{Qwen3-4B} & \textbf{4B-Joint} \\
\midrule
Connectivity & $\uparrow$ & 85.4\% & 90.1\% & 90.2\% & \underline{93.6\%} & \textbf{94.9\%} \\
\cdashline{1-7}[0.4pt/3pt]
Station Grounding & $\uparrow$ & 92.1\% & 92.7\% & 93.8\% & \underline{96.5\%} & \textbf{97.2\%} \\
Distance Plausibility & $\uparrow$ & 72.1\% & 74.0\% & 75.8\% & \underline{84.1\%} & \textbf{87.2\%} \\
\cdashline{1-7}[0.4pt/3pt]
Line Overlap & $\uparrow$ & 0.645 & 0.648 & 0.678 & \underline{0.699} & \textbf{0.706} \\
Station Sequence Overlap & $\uparrow$ & 0.638 & 0.641 & 0.678 & \underline{0.710} & \textbf{0.723} \\
Route Exact Match & $\uparrow$ & 37.3\% & 38.0\% & 42.9\% & \underline{49.1\%} & \textbf{51.8\%} \\
\cdashline{1-7}[0.4pt/3pt]
Estimation Accuracy & $\uparrow$ & 88.4\% & 91.3\% & 90.6\% & \underline{91.7\%} & \textbf{92.3\%} \\
MAPE & $\downarrow$ & 3.04\% & 2.60\% & 2.78\% & \underline{2.35\%} & \textbf{2.10\%} \\
\cdashline{1-7}[0.4pt/3pt]
Preference Compliance & $\uparrow$ & 81.8\% & 81.5\% & 88.2\% & \underline{89.6\%} & \textbf{90.0\%} \\
\bottomrule
\end{tabular}
}
\end{table}

\begin{table}[t]
\centering
\caption{GPS-only ablation on our domain-specific models for \textbf{Multi-Route Generation} with 10{,}000 test samples across four cities. All textual cues are removed and only origin--destination GPS coordinates are provided as input. $\uparrow$\,/\,$\downarrow$ indicate higher/lower is better. \textbf{Bold}: best; \underline{underline}: second best.}
\label{tab:gps_task3}
\small
\setlength{\tabcolsep}{3pt}
\renewcommand{\arraystretch}{1.05}
\scalebox{1.05}{
\begin{tabular}{@{}l@{\enspace}c ccccc@{}}
\toprule
\textbf{Metric} & & \textbf{Qwen3-0.6B} & \textbf{Qwen3-1.7B} & \textbf{Qwen3-4B-25} & \textbf{Qwen3-4B} & \textbf{4B-Joint} \\
\midrule
Connectivity & $\uparrow$ & 92.3\% & 95.2\% & 95.0\% & \underline{96.4\%} & \textbf{97.3\%} \\
\cdashline{1-7}[0.4pt/3pt]
Station Grounding & $\uparrow$ & 95.5\% & 95.7\% & 97.0\% & \underline{98.2\%} & \textbf{98.9\%} \\
Distance Plausibility & $\uparrow$ & 81.0\% & 81.2\% & 84.6\% & \underline{90.1\%} & \textbf{91.8\%} \\
\cdashline{1-7}[0.4pt/3pt]
Line Overlap & $\uparrow$ & 0.751 & 0.753 & 0.750 & \underline{0.774} & \textbf{0.780} \\
Station Sequence Overlap & $\uparrow$ & 0.746 & 0.746 & 0.754 & \underline{0.786} & \textbf{0.796} \\
Route Exact Match & $\uparrow$ & 52.3\% & 52.7\% & 57.4\% & \underline{63.5\%} & \textbf{66.1\%} \\
\cdashline{1-7}[0.4pt/3pt]
Estimation Accuracy & $\uparrow$ & 93.7\% & 95.7\% & 96.1\% & \underline{96.9\%} & \textbf{97.8\%} \\
MAPE & $\downarrow$ & 2.79\% & 2.06\% & 2.30\% & \underline{1.83\%} & \textbf{1.55\%} \\
\cdashline{1-7}[0.4pt/3pt]
Route Diversity & $\uparrow$ & 0.512 & 0.533 & 0.534 & \underline{0.544} & \textbf{0.545} \\
\bottomrule
\end{tabular}
}
\end{table}

\begin{table}[t]
\centering
\caption{Temporal robustness of Qwen3-4B on \textbf{Optimal Route Generation}, evaluated without retraining over a week approximately three months after the official evaluation period, with 10{,}000 samples per day. The last row repeats the Qwen3-4B results on the official test set from Table~\ref{tab:task1_results}. Column abbreviations follow Appendix~\ref{app:metrics}, and $\uparrow$ indicates higher is better.}
\label{tab:temporal}
\small
\setlength{\tabcolsep}{6.5pt}
\renewcommand{\arraystretch}{1.05}
\scalebox{1.05}{
\begin{tabular}{@{}l ccccccc@{}}
\toprule
\textbf{Day} & \textbf{Conn} $\uparrow$ & \textbf{SG} $\uparrow$ & \textbf{DP} $\uparrow$ & \textbf{LO} $\uparrow$ & \textbf{SSO} $\uparrow$ & \textbf{REM} $\uparrow$ & \textbf{EA} $\uparrow$ \\
\midrule
Monday & 97.1\% & 98.4\% & 91.3\% & 0.813 & 0.826 & 69.6\% & 98.5\% \\
Tuesday & 97.1\% & 98.6\% & 91.2\% & 0.827 & 0.838 & 70.2\% & 98.5\% \\
Wednesday & 97.6\% & 98.6\% & 91.6\% & 0.824 & 0.835 & 70.8\% & 98.4\% \\
Thursday & 96.9\% & 98.5\% & 91.7\% & 0.821 & 0.832 & 70.2\% & 98.5\% \\
Friday & 96.8\% & 98.7\% & 91.2\% & 0.820 & 0.833 & 70.0\% & 98.5\% \\
Saturday & 97.4\% & 98.5\% & 91.1\% & 0.818 & 0.829 & 69.8\% & 98.4\% \\
Sunday & 97.1\% & 98.5\% & 91.3\% & 0.816 & 0.826 & 69.1\% & 98.5\% \\
\midrule
Weekly average & 97.1\% & 98.5\% & 91.3\% & 0.820 & 0.831 & 69.9\% & 98.5\% \\
Official test set & 97.0\% & 98.5\% & 91.0\% & 0.828 & 0.838 & 71.0\% & 98.5\% \\
\bottomrule
\end{tabular}
}
\end{table}

\begin{table}[t]
\centering
\caption{Temporal robustness by request time. The 70{,}000 queries of Table~\ref{tab:temporal} are grouped into four six-hour periods of the day, and each row aggregates the queries issued within that period. Column abbreviations follow Appendix~\ref{app:metrics}, and $\uparrow$ indicates higher is better.}
\label{tab:temporal_period}
\small
\setlength{\tabcolsep}{6.5pt}
\renewcommand{\arraystretch}{1.05}
\scalebox{1.05}{
\begin{tabular}{@{}l ccccccc@{}}
\toprule
\textbf{Period} & \textbf{Conn} $\uparrow$ & \textbf{SG} $\uparrow$ & \textbf{DP} $\uparrow$ & \textbf{LO} $\uparrow$ & \textbf{SSO} $\uparrow$ & \textbf{REM} $\uparrow$ & \textbf{EA} $\uparrow$ \\
\midrule
00--06 & 97.3\% & 98.4\% & 91.1\% & 0.819 & 0.831 & 69.9\% & 98.6\% \\
06--12 & 97.0\% & 98.5\% & 91.6\% & 0.822 & 0.833 & 70.1\% & 98.4\% \\
12--18 & 97.2\% & 98.6\% & 91.4\% & 0.816 & 0.828 & 69.6\% & 98.4\% \\
18--24 & 97.0\% & 98.6\% & 91.2\% & 0.822 & 0.834 & 70.1\% & 98.6\% \\
\bottomrule
\end{tabular}
}
\end{table}

\begin{table}[t]
\centering
\caption{Geographic generalization of Qwen3-4B in low-frequency regions on \textbf{Optimal Route Generation}. Origin cells are 500m grids ranked by request frequency over nearly 30 days of real planning traffic, and each cumulative bucket samples 10{,}000 queries from the lowest-frequency cells. \textbf{All} repeats the Qwen3-4B results on the full test set from Table~\ref{tab:task1_results}. $\uparrow$ indicates higher is better. \textbf{Bold}: best. \underline{Underline}: second best.}
\label{tab:sparse}
\small
\setlength{\tabcolsep}{5pt}
\renewcommand{\arraystretch}{1.05}
\scalebox{1.05}{
\begin{tabular}{@{}l@{\enspace}c ccccc@{}}
\toprule
\textbf{Metric} & & \textbf{Lowest 5\%} & \textbf{Lowest 10\%} & \textbf{Lowest 25\%} & \textbf{Lowest 50\%} & \textbf{All} \\
\midrule
Connectivity & $\uparrow$ & 89.8\% & 92.4\% & 95.0\% & \underline{96.6\%} & \textbf{97.0\%} \\
\cdashline{1-7}[0.4pt/3pt]
Station Grounding & $\uparrow$ & 94.9\% & 96.4\% & 97.8\% & \underline{98.0\%} & \textbf{98.5\%} \\
Distance Plausibility & $\uparrow$ & 80.6\% & 85.1\% & 88.5\% & \underline{90.1\%} & \textbf{91.0\%} \\
\cdashline{1-7}[0.4pt/3pt]
Line Overlap & $\uparrow$ & 0.667 & 0.687 & 0.724 & \underline{0.765} & \textbf{0.828} \\
Station Sequence Overlap & $\uparrow$ & 0.673 & 0.700 & 0.732 & \underline{0.774} & \textbf{0.838} \\
Route Exact Match & $\uparrow$ & 38.2\% & 45.2\% & 52.9\% & \underline{60.9\%} & \textbf{71.0\%} \\
\cdashline{1-7}[0.4pt/3pt]
Estimation Accuracy & $\uparrow$ & 97.6\% & 97.9\% & 98.0\% & \underline{98.2\%} & \textbf{98.5\%} \\
\bottomrule
\end{tabular}
}
\end{table}

\begin{table}[t]
\centering
\caption{SFT-stage city transfer on \textbf{Optimal Route Generation}, evaluated on the Beijing test set with 10{,}000 samples. Both models use four-city CPT. The held-out model excludes Beijing from SFT but keeps it in the CPT corpus. Standard SFT repeats the Four-City results from Table~\ref{tab:single_city}. $\uparrow$ indicates higher is better.}
\label{tab:beijing_holdout}
\small
\setlength{\tabcolsep}{6pt}
\renewcommand{\arraystretch}{1.05}
\scalebox{1.05}{
\begin{tabular}{@{}l@{\enspace}c cc@{}}
\toprule
\textbf{Metric} & & \textbf{Standard SFT} & \textbf{Beijing-held-out SFT} \\
\midrule
Connectivity & $\uparrow$ & 95.5\% & 95.1\% \\
\cdashline{1-4}[0.4pt/3pt]
Station Grounding & $\uparrow$ & 97.8\% & 97.4\% \\
Distance Plausibility & $\uparrow$ & 88.8\% & 88.6\% \\
\cdashline{1-4}[0.4pt/3pt]
Line Overlap & $\uparrow$ & 0.845 & 0.822 \\
Station Sequence Overlap & $\uparrow$ & 0.839 & 0.807 \\
Route Exact Match & $\uparrow$ & 70.5\% & 64.5\% \\
\cdashline{1-4}[0.4pt/3pt]
Estimation Accuracy & $\uparrow$ & 97.6\% & 97.4\% \\
\bottomrule
\end{tabular}
}
\end{table}

\begin{table}[h]
\centering
\caption{Tool-augmented evaluation on \textbf{Optimal Route Generation}, using the same six models, 1{,}000 test queries, and output format as Table~\ref{tab:llm_comparison}, with only boarding and alighting stations per leg. Candidates from the Amap transit routing API are supplied in context. Estimation Accuracy and MAPE are omitted because numeric fields are inherited from the API.}
\label{tab:rag_comparison}
\small
\setlength{\tabcolsep}{3pt}
\renewcommand{\arraystretch}{1.05}
\scalebox{1.05}{
\begin{tabular}{@{}l@{\enspace}c cccccc@{}}
\toprule
\textbf{Metric} & & \textbf{GPT-5.4} & \textbf{Deepseek-V4} & \textbf{Gemini-3.1} & \textbf{Claude-4.6} & \textbf{Qwen-3.6} & \textbf{Doubao} \\
\midrule
Connectivity & $\uparrow$ & 97.3\% & \underline{98.0\%} & \textbf{98.3\%} & 96.5\% & 96.7\% & 97.8\% \\
\cdashline{1-8}[0.4pt/3pt]
Station Grounding & $\uparrow$ & 97.2\% & \textbf{100.0\%} & 99.5\% & 99.5\% & 98.7\% & \textbf{100.0\%} \\
Distance Plausibility & $\uparrow$ & 95.1\% & 98.5\% & \underline{98.7\%} & 97.4\% & 97.9\% & \textbf{99.2\%} \\
\cdashline{1-8}[0.4pt/3pt]
Line Overlap & $\uparrow$ & 0.820 & \textbf{0.848} & 0.827 & 0.823 & 0.813 & \underline{0.834} \\
Station Sequence Overlap & $\uparrow$ & 0.810 & \textbf{0.834} & 0.825 & 0.812 & 0.813 & \underline{0.831} \\
Route Exact Match & $\uparrow$ & 71.7\% & \underline{74.2\%} & 73.3\% & 72.1\% & 72.0\% & \textbf{74.4\%} \\
\bottomrule
\end{tabular}
}
\end{table}

\subsection{Data Scaling and GPS-only Ablation on Other Tasks}
\label{app:scaling_additional}

Tables~\ref{tab:scaling_task2} and~\ref{tab:scaling_task3} extend the data scaling analysis from Section~\ref{sec:experiments} to Preference-Aware Planning and Multi-Route Generation. Both tasks exhibit the same monotonic improvement across the main metrics as CPT data volume increases, confirming that the learning hierarchy observed on Optimal Route Generation generalizes to preference-conditioned and multi-route settings. Task-specific metrics also improve overall, with Preference Compliance rising from 87.3\% to 89.8\% on Preference-Aware Planning and Route Diversity from 0.507 to 0.545 on Multi-Route Generation.

Tables~\ref{tab:gps_task2} and~\ref{tab:gps_task3} report the GPS-only ablation results on the same two tasks. Consistent with the findings in Section~\ref{sec:experiments}, all domain-specific models maintain strong performance when textual cues are removed. The 4B-Joint model achieves the best GPS-only results across nearly all metrics on both tasks, with Route Exact Match reaching 51.8\% on Preference-Aware Planning and 66.1\% on Multi-Route Generation. Model size scaling follows the same pattern observed on Optimal Route Generation, with larger models generally outperforming smaller ones under GPS-only input.

\subsection{Temporal Robustness}
\label{app:temporal}

For TransitLM, maintaining route-generation quality over time without frequent retraining is important for practical use. To assess temporal robustness, we reuse the Qwen3-4B model from Table~\ref{tab:task1_results} without retraining and evaluate 70{,}000 Optimal Route Generation queries collected over one week approximately three months after the official evaluation period, with 10{,}000 queries per day. The set follows the official benchmark protocol, including the OD de-duplication in Section~\ref{sec:collection}, with no directed origin--destination pair repeated across days or shared with the pre-training corpus. Table~\ref{tab:temporal} reports daily results, while Table~\ref{tab:temporal_period} groups the same queries into four six-hour periods by request time.

The main temporal change is a modest reduction in exact reference matching, rather than a comparable loss of structural validity or route-specific numeric accuracy. Route Exact Match averages 69.9\%, only 1.1 percentage points below the official benchmark, with a maximum difference of 1.7 percentage points across days. Connectivity, Station Grounding, and Distance Plausibility remain within one percentage point of their official values on every day, while Estimation Accuracy stays within 0.2 percentage points on exactly matched samples. Across request-time periods, percentage-based metrics vary by at most about 0.5 percentage points, while Line Overlap and Station Sequence Overlap each vary by at most 0.006. This stability on directed OD pairs absent from CPT supports reuse of learned routing knowledge across distinct later queries.

These results demonstrate temporal robustness across the evaluated three-month shift and the tested days and request-time periods. Route-generation quality is therefore not confined to the original evaluation period, despite a modest reduction in exact matching. This supports continued model reuse and suggests a reduced need for frequent retraining under the evaluated conditions.

\subsection{Geographic Robustness}
\label{app:geographic}

TransitLM learns transit knowledge from a finite training corpus, while planning requests are geographically uneven. Aggregate performance may therefore conceal weaknesses outside frequently requested locations. We test geographic generalization in low-frequency origin regions and with Beijing excluded from SFT, examining the effects of sparse request coverage and absent target-city task supervision.

For the sparse-region analysis, we rank 500m origin cells by request frequency using about 600 million real planning requests collected over nearly 30 days, and sample 10{,}000 queries from each cumulative bucket of the lowest-frequency 5\%, 10\%, 25\%, and 50\% of cells. The lowest 5\% is the strictest setting. Table~\ref{tab:sparse} shows that structural and spatial validity deteriorate less than exact reference matching. In the sparsest bucket, Connectivity, Station Grounding, and Distance Plausibility decrease to 89.8\%, 94.9\%, and 80.6\%, respectively, whereas Route Exact Match falls from 71.0\% to 38.2\%. Among connected outputs, line and station overlaps of 0.667 and 0.673 indicate substantial reference-route similarity despite reduced exact matching. One explanation is that shared network structure supports planning across queries, while reproducing local route choices benefits from denser examples. This is consistent with the data-scaling results, where smaller CPT data volumes already support high connectivity, while exact matching improves with additional data. Numeric estimates also remain accurate on exactly matched samples, with Estimation Accuracy at least 97.6\%.

To assess SFT-stage transfer, we retain four-city CPT but exclude Beijing from SFT and evaluate on the Beijing test set of 10{,}000 samples. Structural validity is largely preserved in Table~\ref{tab:beijing_holdout}, with Connectivity, Station Grounding, and Distance Plausibility changing by at most 0.5 percentage points, whereas Route Exact Match falls from 70.5\% to 64.5\%. Estimation Accuracy is also nearly unchanged on exactly matched samples. This contrast suggests that CPT-derived transit knowledge remains usable after SFT on other cities, while Beijing-specific supervision further improves reference-route selection.

Together, these experiments support geographic generalization within the transit networks covered by CPT. Connectivity and spatial grounding extend beyond dense query coverage and target-city SFT, while exact reference matching remains more dependent on geographic coverage and local supervision.

\subsection{Tool-Augmented Evaluation of General-Purpose LLMs}
\label{app:rag_comparison}

General-purpose LLMs can obtain transit information through external routing services, so their standalone results do not establish their performance with tool access. We evaluate the gains from routing-engine candidates for all six models in Table~\ref{tab:llm_comparison} under the same queries and output requirements, so that each result can be compared with the standalone performance of the same model. Each LLM acts as an agent that invokes the Amap transit routing API and receives the returned candidates in context. We retain the same 1{,}000 queries across four cities and require only boarding and alighting stations per leg. Connectivity is measured on the generated output, not inferred from candidate validity. Distance, time, and fare are inherited from the API and may reflect conditions different from the static benchmark labels, so Estimation Accuracy and MAPE are omitted.

Table~\ref{tab:rag_comparison} shows large gains across all six models, including those with weak standalone performance. Route Exact Match increases by 33.1--56.7 percentage points, reaching 71.7--74.4\%, while Connectivity reaches 96.5--98.3\%. The gap between the highest and lowest Route Exact Match also contracts from 24.8 to 2.7 percentage points. This convergence is consistent with external candidates supplying route-specific information, reducing reliance on the transit knowledge encoded in each model. The improvement therefore reflects the benefit of tool support, rather than stronger unaided route-generation capability.

These results establish the benefit of tool augmentation for the six general-purpose models under simplified output requirements. The direct comparison with TransitLM is provided separately in Table~\ref{tab:fair_sequence}, using 10{,}000 queries and complete intermediate station sequences to assess route quality with and without external engine access.

\subsection{Quality of Non-Exact Routes}
\label{app:nonexact}

\begin{table}[t]
\centering
\caption{Non-exact connected outputs on \textbf{Optimal Route Generation}, evaluated on model-specific subsets. Candidate Exact Match measures exact agreement with any engine candidate for the same query. Label IoU compares with the reference, while Best IoU is the maximum overlap with any candidate. Qwen3-0.6B through 4B-Joint are TransitLM variants. GPT-5.4 has no tool access. \textbf{Bold} and \underline{underline} mark the highest and second-highest values.}
\label{tab:nonexact}
\small
\setlength{\tabcolsep}{4pt}
\renewcommand{\arraystretch}{1.05}
\scalebox{1.05}{
\begin{tabular}{@{}l@{\enspace}c ccccc@{}}
\toprule
\textbf{Metric} & & \textbf{Qwen3-0.6B} & \textbf{Qwen3-1.7B} & \textbf{Qwen3-4B} & \textbf{4B-Joint} & \textbf{GPT-5.4} \\
\midrule
Candidate Exact Match & $\uparrow$ & 23.6\% & 24.3\% & \underline{34.5\%} & \textbf{39.6\%} & 20.3\% \\
\cdashline{1-7}[0.4pt/3pt]
Station Grounding & $\uparrow$ & 91.5\% & 91.6\% & \underline{96.4\%} & \textbf{97.7\%} & 55.1\% \\
Distance Plausibility & $\uparrow$ & 67.6\% & 66.2\% & \underline{79.3\%} & \textbf{84.6\%} & 47.8\% \\
\cdashline{1-7}[0.4pt/3pt]
Label Line IoU & $\uparrow$ & \underline{0.534} & \textbf{0.545} & 0.452 & 0.419 & 0.512 \\
Label Station IoU & $\uparrow$ & \underline{0.485} & \textbf{0.493} & 0.441 & 0.419 & 0.000 \\
\cdashline{1-7}[0.4pt/3pt]
Best Line IoU & $\uparrow$ & \underline{0.842} & \textbf{0.856} & 0.832 & 0.825 & 0.566 \\
Best Station IoU & $\uparrow$ & 0.735 & 0.746 & \underline{0.771} & \textbf{0.781} & 0.540 \\
\bottomrule
\end{tabular}
}
\end{table}

Many origin--destination pairs admit several reasonable routes, so disagreement with a single reference need not imply a planning failure. We evaluate whether connected, non-exact outputs remain plausible alternatives using Station Grounding, Distance Plausibility, and overlap with the reference and the candidate routes returned by the routing engine for the same query. Table~\ref{tab:nonexact} reports each model on its own non-exact connected subset. These conditional results should not be directly compared with the full-test-set results in Tables~\ref{tab:task1_results} and~\ref{tab:llm_comparison}.

Reference mismatch coexists with spatial plausibility and substantial overlap with alternatives. Across the four TransitLM variants, Station Grounding ranges from 91.5\% to 97.7\% and Distance Plausibility from 66.2\% to 84.6\%. Both line and station overlap are markedly higher against the best matching candidate than against the reference. This pattern suggests that some mismatches reflect alternative route choices rather than arbitrary errors. The weaker grounding and candidate overlap of GPT-5.4 illustrate why connectivity alone is insufficient to establish route quality. Grounding and distance checks provide evidence beyond candidate similarity by assessing alignment with the requested endpoints and plausible access distances.

Candidate matching reveals a further distinction between route similarity and exact reproduction. Only 23.6\% to 39.6\% of these TransitLM outputs exactly match a returned candidate, despite Best Line IoU of 0.825 to 0.856 and Best Station IoU of 0.735 to 0.781. Since TransitLM receives no candidates at inference time, this pattern is consistent with applying learned routing knowledge beyond exact candidate reproduction. Candidate mismatch alone could reflect errors, so grounding, distance plausibility, and candidate overlap are essential complementary evidence for this interpretation.

Overall, non-exact outputs can remain plausible alternatives, making exact reference matching an incomplete measure of route quality. The results also provide evidence beyond simple reproduction of returned candidates, without establishing independence from engine supervision. The quality assessment relies on automated checks and a finite candidate set, with large-scale human validation still needed to establish practical acceptability.

\section{Limitations and Future Work}
\label{app:limitations}

TransitLM internalizes routing knowledge in model parameters and generates routes from OD queries and natural-language constraints without external map or routing-engine access at inference time. This makes it complementary to production routing engines, not a replacement for them. Generated routes are not guaranteed to be valid, which limits deployment in settings requiring strict reliability. Practical deployment would therefore need route validation and fallback handling for outputs that fail reliability checks.

The dataset covers four Chinese cities and 120{,}845 stations, with all text in Chinese. Transfer to networks with different topologies, transfer conventions, or languages remains unverified. Dedicated station tokens make vocabulary size grow linearly with geographic coverage, increasing memory and computational costs. More compact or hierarchical station representations are therefore an important direction for geographic expansion.

The model represents a static network snapshot. The temporal stability observed in Appendix~\ref{app:temporal} does not imply automatic adaptation to congestion, service suspensions, or new stations and lines. Incorporating network changes currently requires updating training data and retraining the model. Adding previously unseen stations also requires extending the station vocabulary, not merely supplying updated service information. Future work should reduce this update cost and explore hybrid extensions that retrieve dynamic information at inference time.

Near-term priorities are improving reliability on difficult OD queries, station grounding, and cost consistency, alongside preference compliance and multi-route diversity. As Appendix~\ref{app:nonexact} shows, reference mismatch need not imply failure, so these improvements should be assessed beyond exact matching. A longer-term direction is a transit foundation model integrating question answering, context-aware recommendation~\cite{yan2026inttravel,guo2026towards,wang2026genmrp,chen2026scasrec,yu2025dsfnet}, and route planning, supported by broader task data and appropriate privacy safeguards.

\section{Ethics and Privacy}
\label{app:ethics}

TransitLM is constructed from transit planning logs provided by Amap. Its planning sessions are isolated OD requests rather than continuous GPS trajectories such as those in T-Drive~\cite{yuan2010tdrive} and GeoLife~\cite{zheng2010geolife}. The single-day sampling restriction applies to CPT planning sessions, while benchmark queries are collected on later days. The released data include OD coordinates, POI information, route descriptions, route-selection labels, and numeric estimates. User identifiers, demographic attributes, and device fingerprints are excluded, and no user-level linkage keys or longitudinal user histories are provided. These measures reduce information available for linking requests and reconstructing mobility routines, but location information can remain sensitive and de-identification does not guarantee anonymity~\cite{de2013unique}.

The current model is intended for structured transit route generation, not user profiling or longitudinal behavior modeling. Route-selection labels capture planning choices without providing individual histories. The method may broaden access to route-planning capabilities without dependence on external routing services. Incorrect routes or numeric estimates could mislead travelers, cause delays, or incur additional costs. Practical deployment would therefore require route validation, reliability checks, and fallback handling. The benchmark evaluates generation quality rather than a safeguarded deployment system. Extensions involving user-specific context would require separate privacy assessment and safeguards.

\section{Qualitative Examples}
\label{app:case_study}

\begin{figure}[t]
\centering
\includegraphics[width=0.96\textwidth]{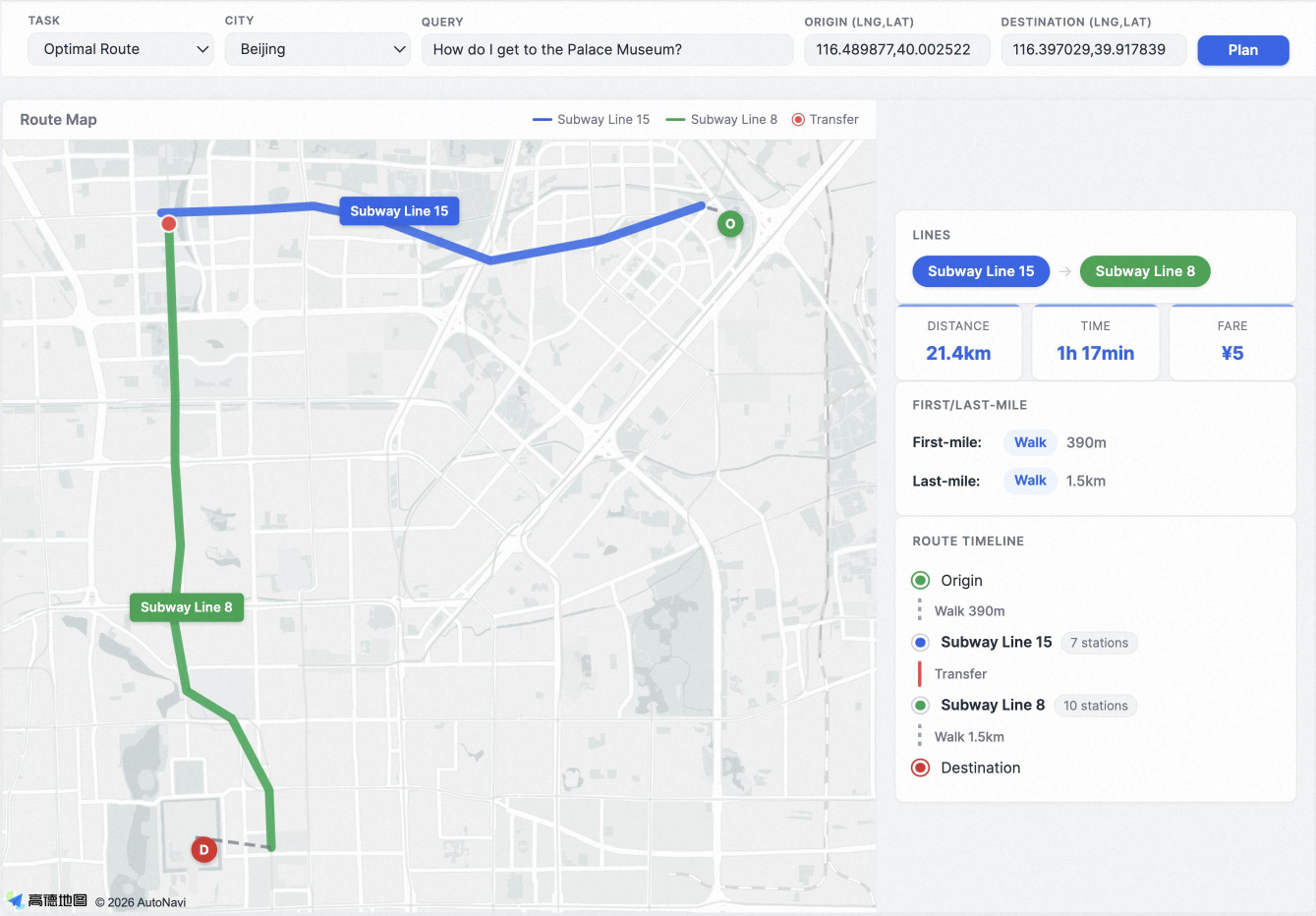}
\caption{Optimal Route Generation example from the 4B-Joint model in Beijing. Given a natural-language query and origin--destination coordinates, the model generates a two-segment subway route with one transfer, together with distance, time, fare, and first/last-mile estimates. Left: route plotted from station-level GPS coordinates in the output. Right: structured generation.}
\label{fig:demo_optimal}
\end{figure}

\begin{figure}[t]
\centering
\includegraphics[width=0.96\textwidth]{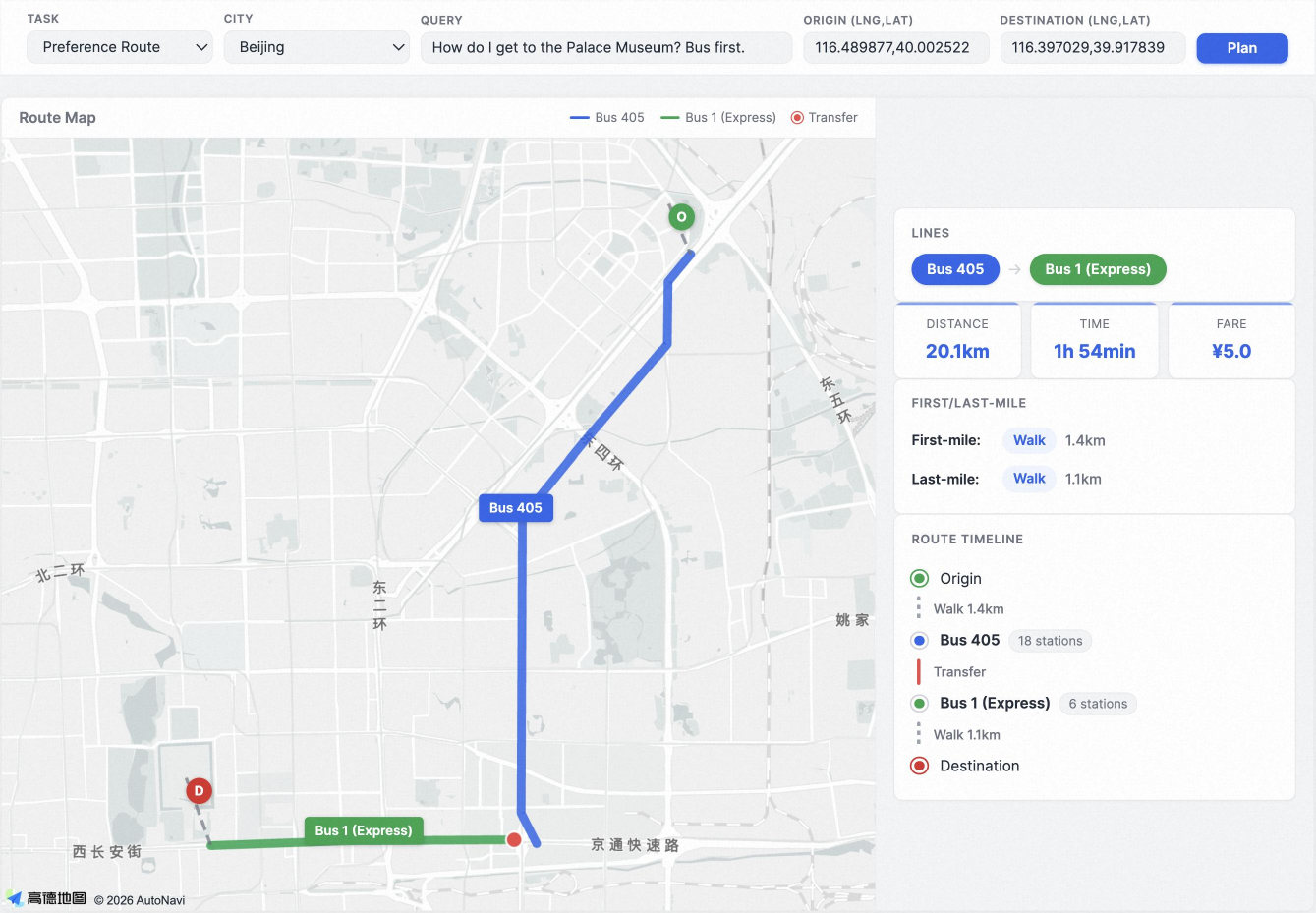}
\caption{Preference-Aware Planning example on the same OD pair with an added ``bus first'' constraint. The model avoids all subway segments and generates a bus-only route via Bus~405 and Bus~1~Express, demonstrating preference compliance.}
\label{fig:demo_preference}
\end{figure}

\begin{figure}[t]
\centering
\includegraphics[width=0.96\textwidth]{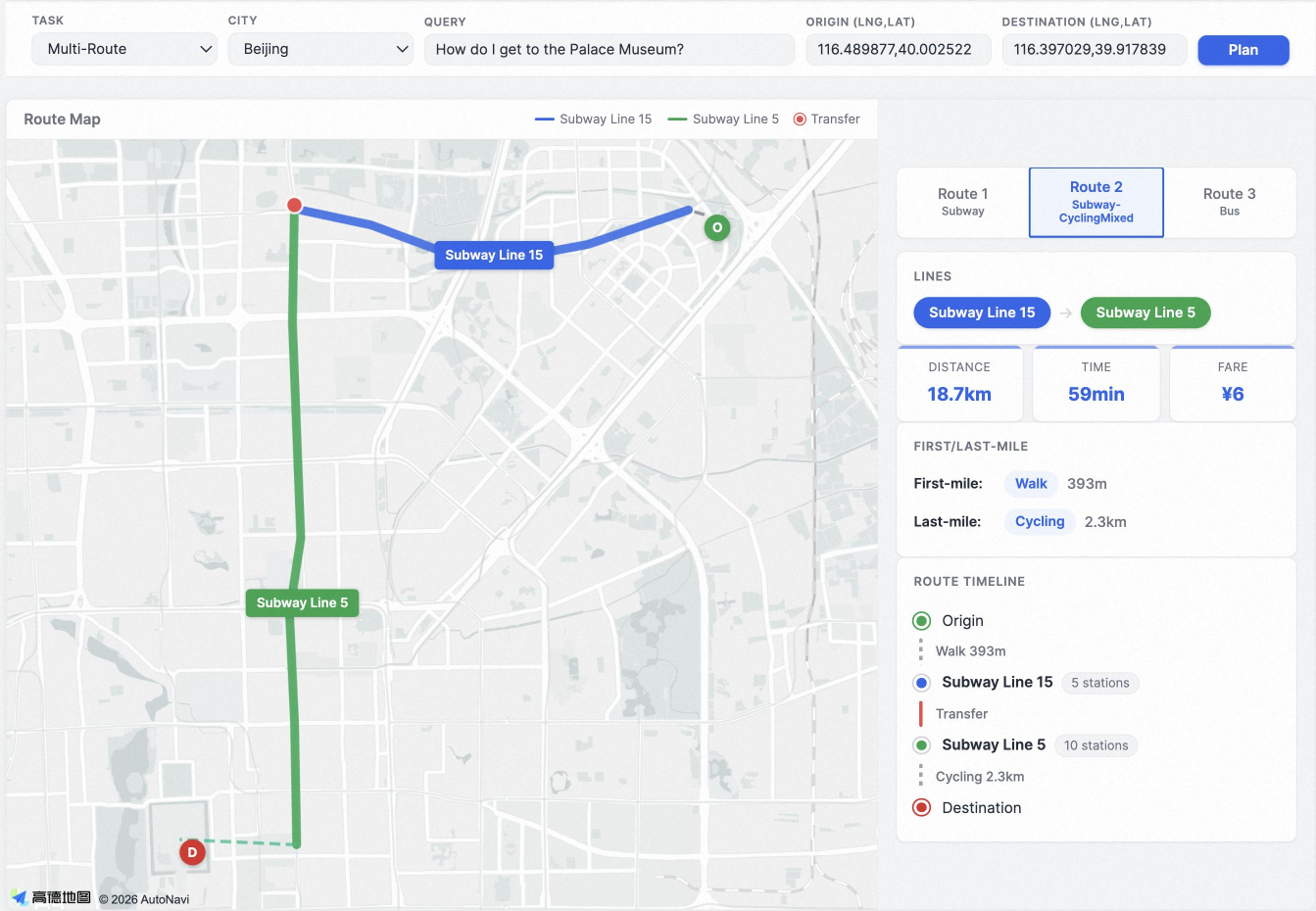}
\caption{Multi-Route Generation example on the same OD pair. The model produces three alternatives spanning subway, subway with cycling, and bus. Route~2 is visualized due to space constraints.}
\label{fig:demo_multi}
\end{figure}

\begin{figure}[t]
\centering
\includegraphics[width=0.96\textwidth]{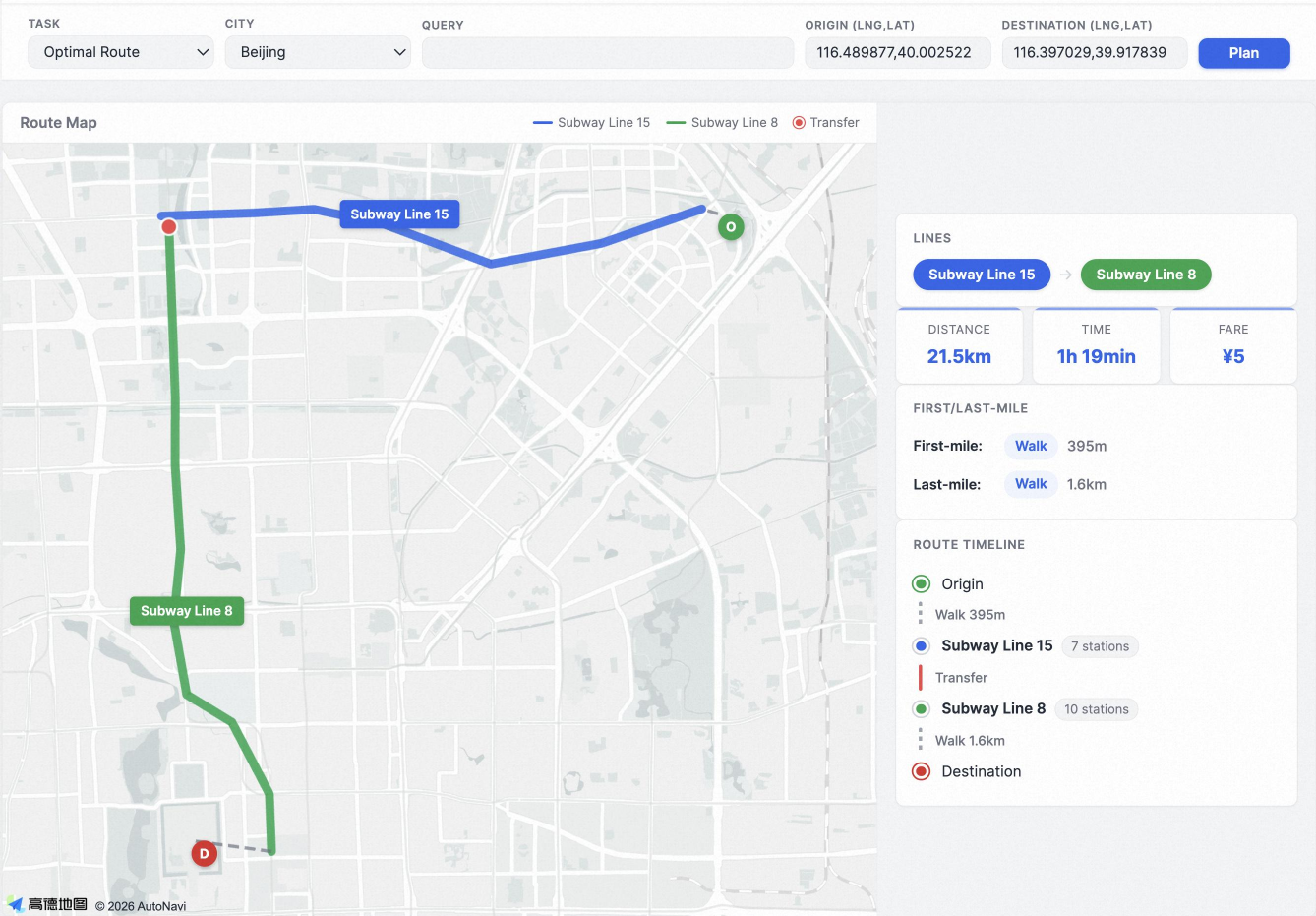}
\caption{GPS-only Optimal Route Generation on the same OD pair with the textual query removed. The generated route is nearly identical to Figure~\ref{fig:demo_optimal}, illustrating robustness to the removal of textual cues.}
\label{fig:demo_gps}
\end{figure}

Figures~\ref{fig:demo_optimal}--\ref{fig:demo_gps} present qualitative outputs from the Qwen3-4B-Joint model on a fixed origin--destination pair in Beijing, with identical GPS coordinates across all four examples enabling direct comparison across tasks and input modalities. The left panel plots each route from the station-level GPS coordinates in the model output, and the right panel shows the structured generation, including line sequences, transfer points, and numeric estimates. All outputs are produced by autoregressive text generation without an external map-matching engine or routing algorithm.

Figure~\ref{fig:demo_optimal} shows the Optimal Route Generation result. Given a natural-language query and origin--destination coordinates, the model generates a two-segment subway route with a single transfer, covering 21.4\,km in 1\,h\,17\,min at \textyen5. The generated station sequence forms a spatially coherent path, and the model selects walking as the first/last-mile access mode with plausible distances.

Figure~\ref{fig:demo_preference} illustrates Preference-Aware Planning on the same OD pair with an added ``bus first'' constraint. The model switches entirely from subway to bus lines, producing a route via Bus~405 and Bus~1~Express that avoids all subway segments. This example illustrates route selection conditioned on the requested preference.

Figure~\ref{fig:demo_multi} presents Multi-Route Generation, where the model produces three distinct alternatives for the same query. The three routes span different transport modalities, with Route~1 using subway only, Route~2 combining subway with cycling for last-mile access, and Route~3 relying on bus. Due to space constraints, only Route~2 is visualized. These outputs illustrate diverse planning choices for the same query.

Figure~\ref{fig:demo_gps} repeats the Optimal Route Generation task with the textual query removed, retaining only raw GPS coordinates as input. The model produces a route nearly identical to Figure~\ref{fig:demo_optimal}, selecting the same subway lines and transfer station with comparable distance and travel time estimates. This example is consistent with the GPS-only robustness reported in Section~\ref{sec:experiments}.

\newpage

\clearpage

\end{document}